\documentclass[journal]{IEEEtran}
\usepackage{graphicx}  
\usepackage{booktabs}  
\usepackage{subfigure}  
\usepackage{algorithm}  
\usepackage{algorithmic}  
\usepackage{multirow}  
\usepackage{amsfonts}  
\usepackage{color,xcolor}  
\usepackage{amssymb}  
\usepackage{amsfonts}  
\usepackage{url}  
\usepackage{ulem}  
\usepackage{amsmath}
\usepackage{placeins}
\newcommand{\eg}{e.g.}
\newcommand{\ie}{i.e.}
\renewcommand{\uline}{}

\begin{document}
\title{A Simple and Strong Baseline for Universal Targeted Attacks on Siamese Visual Tracking}
\author{
  Zhenbang Li, Yaya Shi, Jin Gao, Shaoru Wang,\\ Bing Li, Pengpeng Liang, Weiming Hu,~\IEEEmembership{Senior Member,~IEEE}
  \thanks{This work is supported by National Key R\&D Program of China (No. 2018AAA0102802, No. 2018AAA0102803, No. 2018AAA0102800), Natural Science Foundation of China (Grant No. 61972394, 62036011, 61721004, 61906192), the Key Research Program of Frontier Sciences, CAS, Grant No. QYZDJ-SSW-JSC040, and the Science and Technology Service Network Initiative, CAS, Grant No. KFJ-STS-SCYD-317. Jin Gao and Bing Li are also supported in part by the Youth Innovation Promotion Association, CAS. \textit{(Corresponding author: Jin Gao.)}}
  
  \thanks{Z. Li, J. Gao, S. Wang and B. Li are with National Laboratory of Pattern Recognition, Institute of Automation, Chinese Academy of Sciences, Beijing, 100190, China, and also with School of Artificial Intelligence, University of Chinese Academy of Sciences, Beijing, 100190, China (e-mail: \{zhenbang.li, jin.gao, bli\}@nlpr.ia.ac.cn, wangshaoru2018@ia.ac.cn).}
  \thanks{Y. Shi is with University of Science and Technology of China, Anhui, 230026, China (e-mail: shiyaya@mail.ustc.edu.cn).}
  \thanks{P. Liang is with Zhengzhou University, Henan, 450001, China (e-mail: liangpcs@gmail.com).}
  \thanks{W. Hu is with CAS Center for Excellence in Brain Science and Intelligence Technology, National Laboratory of Pattern Recognition, Institute of Automation, Chinese Academy of Sciences, Beijing, 100190, China, and also with School of Artificial Intelligence, University of Chinese Academy of Sciences, Beijing, 100190, China (e-mail: wmhu@nlpr.ia.ac.cn).}
}

\markboth{Journal of \LaTeX\ Class Files,~Vol.~14, No.~8, August~2015}
{Shell \MakeLowercase{\textit{et al.}}: Bare Demo of IEEEtran.cls for IEEE Journals}
\maketitle

\begin{abstract}
Siamese trackers are shown to be vulnerable to adversarial attacks recently. However, the existing attack methods craft the perturbations for each video independently, which comes at a non-negligible computational cost. In this paper, we show the existence of universal perturbations that can enable the targeted attack, e.g., forcing a tracker to follow the ground-truth trajectory with specified offsets, to be video-agnostic and free from inference in a network. Specifically, we attack a tracker by adding a universal translucent perturbation to the template image and adding a \textit{fake target}, i.e., a small universal adversarial patch, into the search images adhering to the predefined trajectory, so that the tracker outputs the location and size of the \textit{fake target} instead of the real target. Our approach allows perturbing a novel video to come at no additional cost except the mere addition operations -- and not require gradient optimization or network inference. Experimental results on several datasets demonstrate that our approach can effectively fool the Siamese trackers in a targeted attack manner. We show that the proposed perturbations are not only universal across videos, but also generalize well across different trackers. Such perturbations are therefore doubly universal, both with respect to the data and the network architectures. We will make our code publicly available.

\end{abstract}

\IEEEpeerreviewmaketitle

\section{Introduction}
\IEEEPARstart{G}{iven} an arbitrary detected or annotated object of interest in the initial video frame, visual object tracking is aimed at {\it recognizing} and {\it localizing} other instances of the same object in subsequent frames. This paradigm of tracking visual objects from a single initial exemplar in the online-tracking phase has been broadly cast as a Siamese network-based one-shot problem recently \cite{DBLP:journals/tcsv/ShanZLZH21,DBLP:journals/tcsv/FanSZYL21,DBLP:journals/tcsv/LiPWK20,9258983}, which is termed as Siamese visual tracking and recognised as highly effective and efficient for visual tracking.

\begin{figure}[t]
  \centering
  \includegraphics[width=0.45\textwidth]{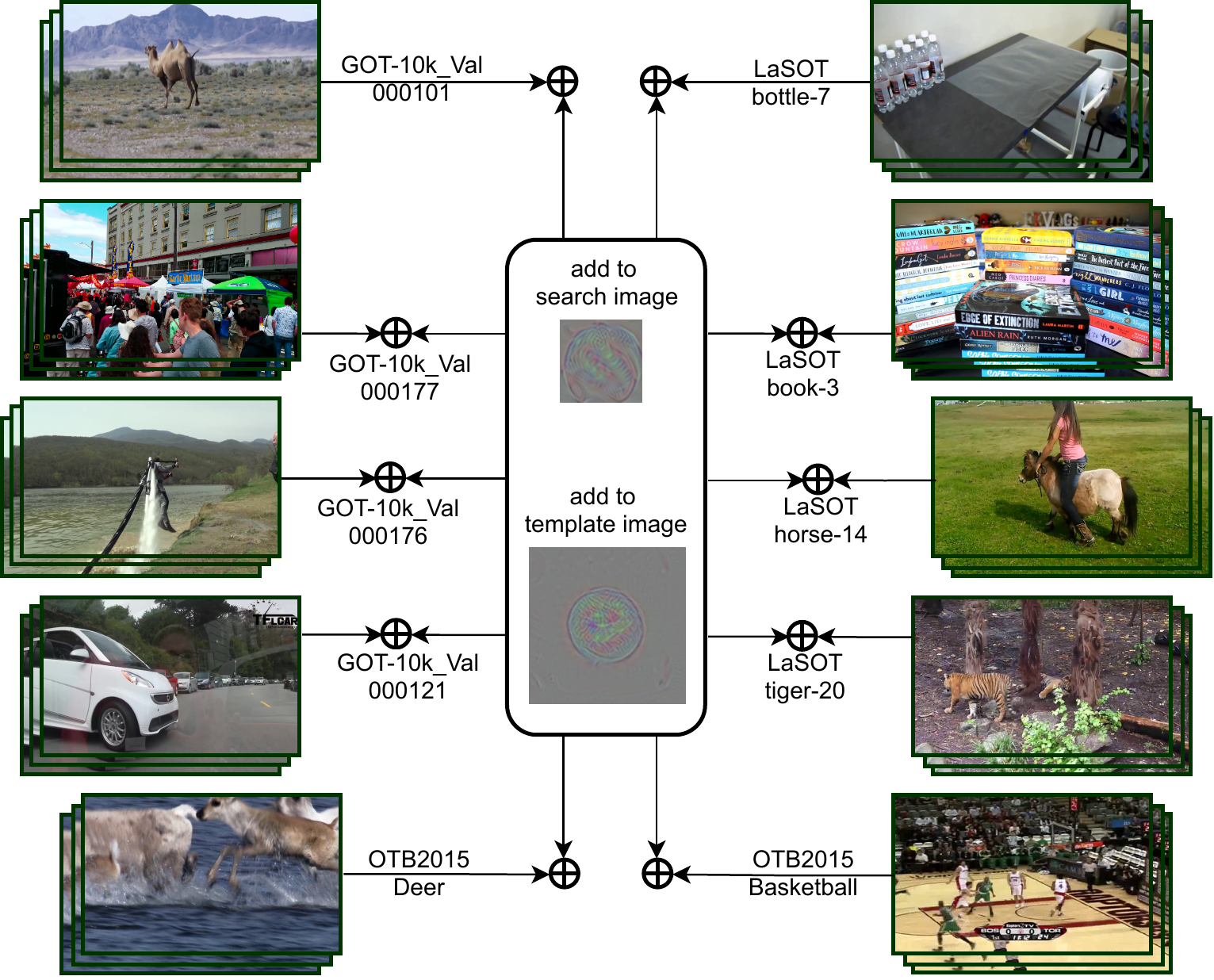}
  \caption{Being universal, our generated perturbations can be conveniently exploited to perturb videos on-the-fly without extra intensive computations except the mere addition operations.}
  \label{fig:UAP}
  \vspace{-4mm}
\end{figure}

Recently, the robustness of Siamese trackers has attracted much attention in the sense of testing their vulnerability to adversarial attacks, and the focus has been directed towards more efficient and low-cost attacks \cite{TTP,FAN,SPARK,chen2020one}. However, these attack methods still need to craft the perturbations for each video independently based on either iterative optimization or adversarial network inference, for which the adequacy of additional computational resources may be not ensured by the computationally intensive tracking systems. 

Universal adversarial perturbations (UAPs) proposed in \cite{UAP} can fool most images from a data distribution in an image-agnostic manner. Being universal, UAPs can be conveniently exploited to perturb unseen data on-the-fly without extra intensive computations. However, no existing work has touched the topic of attacking the Siamese trackers using UAPs, because it is hard to apply existing UAPs to attack Siamese trackers directly. The main reason lies in the fact that, (a) most UAPs are designed for typical neural networks with one image as input while Siamese networks accept both the template and search images, and (b) the goal of existing UAP methods is to disturb unary or binary model outputs for single instance while we need to use universal perturbations to mislead Siamese trackers to follow a specified trajectory.
 
\begin{figure}[t]
  \centering
  \includegraphics[width=0.45\textwidth]{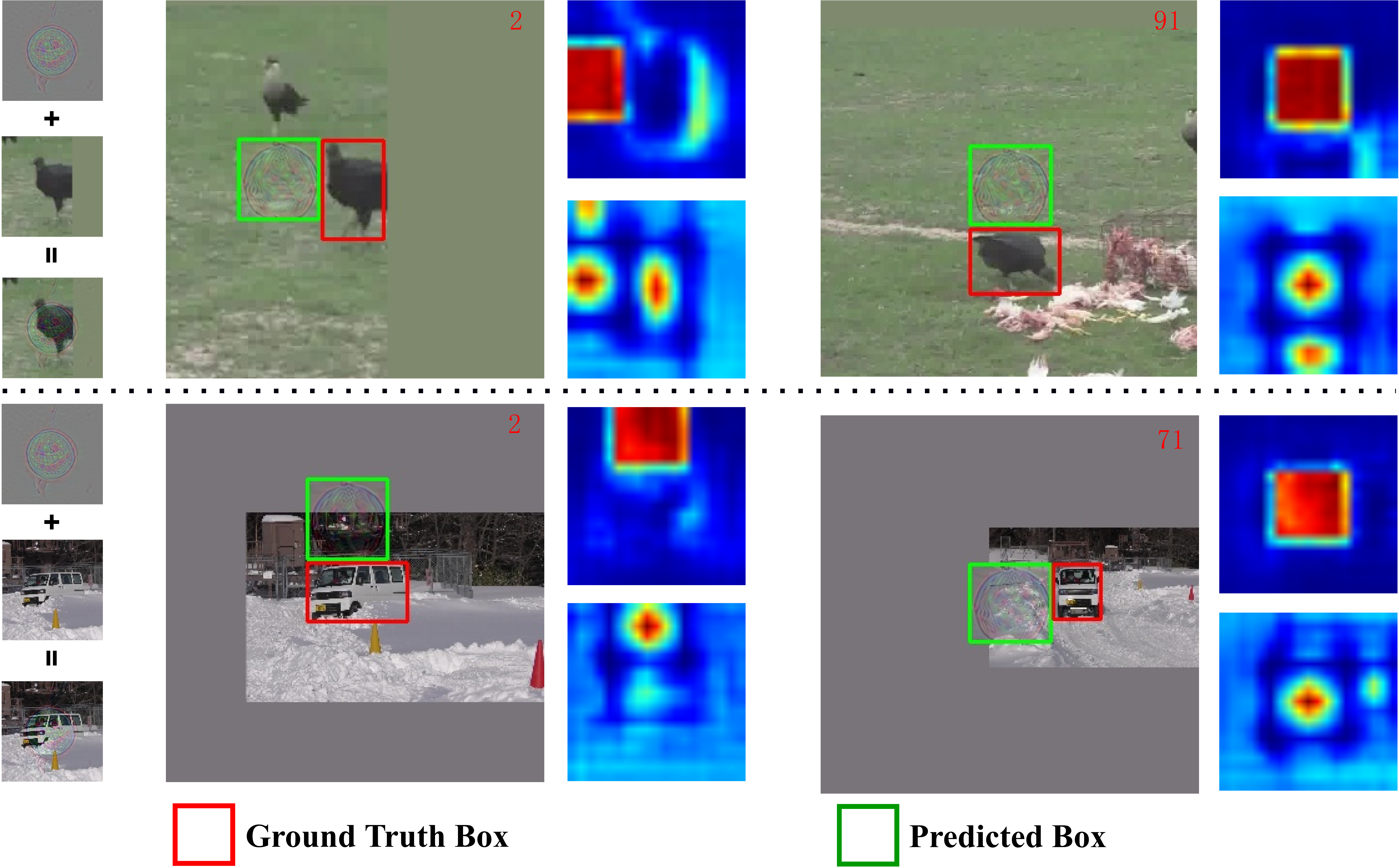}
  \caption{An illustration of our attacks to SiamFC++ on some example tracking sequences from GOT-10k benchmark. Our approach generates the video-agnostic perturbations which can force SiamFC++ to follow a complicated trajectory at virtually no cost. This is realized by off-line training the perturbations so that the tracker mistakenly believes the \textit{fake target} area contains the object to be tracked (see the top heatmap). Moreover, the quality assessment branch in SiamFC++ is also misled to confirm this result (see the bottom heatmap). The \textit{fake target} size is gradually decreased.} 
  \label{fig:1}
  \vspace{-5mm}
\end{figure}
  
In this paper, we make the first attempt in finding the universal perturbations (see Fig.~\ref{fig:UAP}) that fool a state-of-the-art Siamese tracker, \ie, SiamFC++ \cite{SiamFC++}, in a targeted attack manner, which comes at virtually no cost in the online-tracking phase. Specifically, we aim to attack the trackers by adding a universal perturbation to the template image and adding a \textit{fake target}, \ie, a small universal adversarial patch, into the search \uline{images} adhering to the predefined trajectory (as shown in Fig.~\ref{fig:1}), so that the tracker outputs the location and size of the \textit{fake target} instead of the real target. Our generated video-agnostic perturbations allow perturbing a novel video to come at no additional cost except the mere addition operations -- and not require gradient optimization or network inference. Experimental results on OTB2015 \cite{OTB}, GOT-10k \cite{GOT-10k}, LaSOT \cite{LaSOT}, VOT2016 \cite{VOT2016}, VOT2018 \cite{VOT2018} and VOT2019 \cite{VOT2019} demonstrate the effectiveness and efficiency of our approach.

\section{Related Work}

\subsection{Siamese Visual Tracking}

Siamese visual tracking is a fundamental research direction in template matching-based tracking besides the correlation filter-based methods. Both of them are aimed to ``causally'' estimate the positions of a template cropped from the initial video frame in the subsequent frames. Online learning has play an important role in correlation filter-based tracking, and the recent works (\eg,~\cite{AutoTrack, 9376997, 9132673}) have largely \uline{advanced} the research in this area by exploring various online update strategies. However, most Siamese trackers do not exploit the online learning paradigm and innovatively formulate visual tracking as learning cross-correlation similarities between a target template and the candidates in the search region in an end-to-end convolution fashion. Tracking is then performed by locating the object in the search image region based on the highest visual similarity. This paradigm is formulated as a local one-shot detection task.

Recently, some Siamese trackers~\cite{SiamFC++}, \cite{zhang2020ocean, cui2020fully, 9157720}, \cite{SiamRPN,SiamRPN++} have demonstrated a significant performance improvement in visual tracking. In particular, SiamRPN \cite{SiamRPN} consists of one Siamese subnetwork for feature extraction and another region proposal subnetwork including the classification and regression branches separately. \uline{Based on the success of the decomposition of classification and state estimation in SiamRPN,} SiamRPN++ \cite{SiamRPN++} further breaks the restriction of strict translation invariance through a simple yet effective spatial aware sampling strategy and successfully trains a ResNet-driven Siamese tracker with significant performance gains. Apart from this kind of anchor-based methods, an anchor-free tracker SiamFC++ \cite{SiamFC++} is further designed by considering non-ambiguous scoring, prior target scale/ratio distribution, knowledge-free and estimation quality assessment guidelines. In this anchor-free paradigm, many recent works (\eg,~\cite{zhang2020ocean, cui2020fully, 9157720}) are dedicated to more robust and efficient visual tracking based on various anchor-free settings. Ocean \cite{zhang2020ocean} directly predicts the position and scale of target objects in an anchor-free fashion; since each sample position in \uline{ground-truth} boxes is well trained, the tracker is capable of rectifying inexact predictions of target objects during inference. FCOT \cite{cui2020fully} introduces an online regression model generator (RMG) based on the carefully designed anchor-free box regression branch, which enables the tracker to be more effective in handling target deformation during tracking procedure. SiamCAR \cite{9157720} is both anchor and proposal free, and takes one unique response map to predict object location and bounding box directly; this setting significantly reduces the number of hyper-parameters, which keeps the tracker from complicated parameter tuning and makes the tracker significantly simpler, especially in training. In our experiments, we are focused on the anchor-free SiamFC++ tracker, whereas the transferability of our generated adversarial attacks to some other Siamese trackers is also studied. A comprehensive survey of the related trackers is beyond the scope of this paper\uline{;} please refer to \cite{9339950} for a thorough survey.
\vspace{-2mm}

\subsection{Adversarial Attacks}

Adversarial attack \cite{9169672} to image classification is first investigated in \cite{intriguing} with the aim of identifying the vulnerability of modern deep networks to imperceptible perturbations. Recent studies also emerge to investigate the adversarial attacks to other diverse types of tasks such as natural language processing \cite{generating,zhang2020adversarial,morris2020textattack,jin2020bert} and object detection \cite{wei2019transferable}. Some recent research topics which are mostly related to adversarial attack also include digital watermarking \cite{9343885}, 3D face presentation attacks to the face recognition networks \cite{9294085} and adversarial defense for image classifiers \cite{9169672}. Xiong \textit{et al.} \cite{9343885} propose to generate the digital watermarking by slightly modifying the pixel values of video frames to protect video content from unauthorized access. Compared with \cite{9343885}, our purpose of modifying pixel values of video frames is attacking the tracker, instead of detecting the illegal distribution of a digital movie. Jia \textit{et al.} \cite{9294085} propose to generate 3D face artifacts to attack the face recognition \uline{networks}. Compared with \cite{9294085}, our method directly modifies the input of the network instead of changing the input of cameras using 3D face artifacts. Wang \textit{et al.} \cite{9169672} propose the white-box attack/defense methods for image classifiers. Compared with \cite{9169672}, we focus on attacking the object tracking networks instead of the image classification networks. In the following, we mainly introduce some scenarios of possible adversarial attacks which can be categorized along different dimensions.

\textit{White box attacks v.s. Black box attacks\uline{.}} In the white box attack setting \cite{meng2019white}, the adversary has full knowledge of the model including model type, model architecture and values of all parameters and trainable weights. In the black box setting \cite{cheng2018query,li2019nattack,papernot2017practical,li2020projection}, the adversary has limited or no knowledge about the model under attack \cite{kurakin2018adversarial}. In this paper, we focus on the white box \uline{attacks to} Siamese trackers. However, we surprisingly find that \uline{our} perturbations based on the SiamFC++ tracker~\cite{SiamFC++} can perform black box attacks to some other Siamese trackers such as SiamRPN \cite{SiamRPN}, SiamRPN++ \cite{SiamRPN++} and Ocean \cite{zhang2020ocean}.

\textit{Non-universal attacks v.s. Universal attacks\uline{.}} In the setting of non-universal attacks \cite{dai2018adversarial,li2018second,lin2017tactics}, the adversary has to generate one perturbation for every new datapoint, which is time consuming. In the setting of universal attacks \cite{khrulkov2018art,mopuri2018nag,zhang2020understanding,mopuri2018generalizable,chen2018shapeshifter}, only one perturbation is generated and \uline{then} used for \uline{all datapoints in} the dataset. In this paper, we focus on generating universal perturbations for Siamese trackers to perform efficient attacks.

\textit{Imperceptible Perturbations v.s. Adversarial Patch\uline{.}} The imperceptible perturbations most commonly modify each pixel by a small amount and can be found using a number of optimization strategies such as Limited-memory BFGS \cite{intriguing} and PGD \cite{PGD}. Different from the imperceptible perturbations, the adversarial patch is extremely salient to a neural network. The adversarial patch can be placed anywhere into the input image to cause the network to misbehave, and thus is commonly used for universal attacks \cite{patch}. Note that our adversarial patch works in the image domain instead of the network domain. In the network-domain case, the noise is allowed to take any value and is not restricted to the dynamic range of image value as in the image-domain case \cite{karmon2018lavan}.

\textit{Untargeted Attacks v.s. Targeted Attacks\uline{.}} In the case of untargeted attacks, the adversary's goal is to cause the network to predict any incorrect label and whatever the incorrect label is does not matter, \eg, pushing the object location estimation just outside the true search region in visual tracking. Targeted attacks, however, aim to change the network's prediction to some specific target label. In visual tracking, the targeted attacks aim to intentionally drive trackers to output specified object locations following a predefined trajectory.
\vspace{-2mm}

\subsection{Adversarial Attacks in Visual Tracking} \label{attacktracker}
Recently, there are several explorations of the adversarial attacks to the visual tracking task. For example, PAT \cite{PAT} generates physical adversarial textures via white-box attacks to steer the tracker to lock on the texture when a tracked object moves in front of it. However, PAT validates its method by attacking a light deep regression tracker GOTURN \cite{GOTURN}, which has low tracking accuracy on modern benchmarks. In this paper, we aim to attack state-of-the-art Siamese trackers. RTAA \cite{RTAA} takes temporal motion into consideration when generating lightweight perturbations over the estimated tracking results frame-by-frame. \uline{CSA \cite{CSA} is a Siamese tracking attack method called cooling-shrinking attack. CSA can suppress the peak region of the heat map reflecting the target location, which is used to attack the tracker's targeting ability. However, both of these two methods only perform} the untargeted attacks \uline{to} trackers, which is less challenging than the targeted attacks in this paper, as we aim to create arbitrary, complex trajectories at test time. 

Targeted attacks to follow an erroneous path which looks realistic are crucial to deceive the real-world tracking \uline{systems} without raising possible suspicion. SPARK \cite{SPARK} computes incremental perturbations by using information from the past frames to perform targeted attacks \uline{to} Siamese trackers. However, SPARK needs to generate distinct adversarial examples for every search image through heavy iterative schemes, which is time-consuming to attack online tracking in real time. \uline{FAN~\cite{FAN} proposes a fast attack network for attacking SiamFC tracker. To perform untargeted attacks, FAN proposes a drift loss that shifts the tracker's prediction of the target's position. The tracking error accumulates over time until the tracker loses the target completely. To perform targeted attacks, FAN proposes embedding feature loss for increasing the similarity between the features of the adversarial sample and the regions specified by a particular trajectory. However, similar to CSA, FAN also requires running a generative network for each frame to obtain adversarial information, making it difficult to meet the demand for real-time tracking.} The recent real-time attacker TTP \cite{TTP} exclusively uses the template image to generate \uline{a single} temporally-transferable perturbation in a one-shot manner, and then adds it to every search image of the video. However, this method still needs to generate \uline{the perturbation} for each individual video, and its targeted attack setting requires diverse perturbations from several runs of network inference. It is ill-suited to attack a real-world online-tracking system when we can not get access to the limited computational resources. In this paper, however, we propose video-agnostic perturbations which allow perturbing a novel video to come at no additional cost except the mere addition operations.

\section{Method}\label{method}

In this section, we introduce our video-agnostic targeted attack framework for Siamese trackers. We aim to attack the tracker by adding a perturbation to the template image and adding a \textit{fake target}, i.e., an adversarial patch, into the search images adhering to the predefined trajectory, so that the tracker outputs the location and size of the \textit{fake target} instead of the real target. Below, we formalize our targeted attacks to SiamFC++ \cite{SiamFC++}, and then introduce our perturbation strategy.
\vspace{-5mm}
 
\subsection{Problem Definition}\label{problemdefinition}

Let $V=\{I_i\}_1^T$ denote the frames of a video sequence of length $T$. $B^{gt}=\{b^{gt}_i\}_1^T$ are used to represent the target's ground-truth positions in those frames. Visual object tracking aims to predict the positions $B^{pred}=\{b^{pred}_i\}_1^T$ of the target in the subsequent frames given its initial state. In SiamFC++, the tracker first transforms the paired reference frame $I_1$ and annotation $b_1^{gt}$ to get a template image $\textbf z_1$, and transforms the search frame $I_i$ to get the search image $\textbf x_i$ centered at the position estimated in the previous frame. At each time-step, the template image $\textbf z_1$ and the search image $\textbf x_i$ are first passed individually through a shared backbone network, \ie, Siamese network, and then fused using a channel-wise correlation operation:
\begin{equation}
  \text{Feat}_{j}(\mathbf{z}_1, \mathbf{x}_i)=\psi_{j}(\phi(\mathbf{z}_1)) \star \psi_{j}(\phi(\mathbf{x}_i)), j \in\{\mathrm{cls}, \mathrm{reg}\}
\end{equation}
where $\star$ denotes the channel-wise correlation operation, $\phi(\cdot)$ denotes the feature extractor of the Siamese network, $\psi_j(\cdot)$ denotes the layers specific to the classification or regression task, and $j$ denotes the specific task type ($\mathrm{cls}$ denotes the classification task and $\mathrm{reg}$ denotes the regression task). $\psi_{\mathrm{cls}}$ and $\psi_{\mathrm{reg}}$ are both designed as two \uline{convolution} layers for adapting generic features to the feature space specific to the classification/regression task. The fused features then act as input to a head network, which predicts a classification map $\textbf{C}$, a bounding box regression map $\textbf{R}$, and a quality assessment map $\textbf{Q}$ in an anchor-free manner. In short, $\textbf C$ encodes the probability of each spatial position to contain the target, $\textbf R$ regresses the bounding box of the target, and $\textbf Q$ predicts the target state estimation quality. The final bounding box is then generated according to $\textbf{C}$, $\textbf{R}$ and $\textbf{Q}$.

A straight forward way to achieve targeted attacks \uline{to} Siamese trackers is directly using popular attack methods such as FGSM \cite{FGSM} and BIM \cite{DBLP:conf/iclr/KurakinGB17a}:

\begin{figure}[t]
  \centering
  \includegraphics[width=0.45\textwidth]{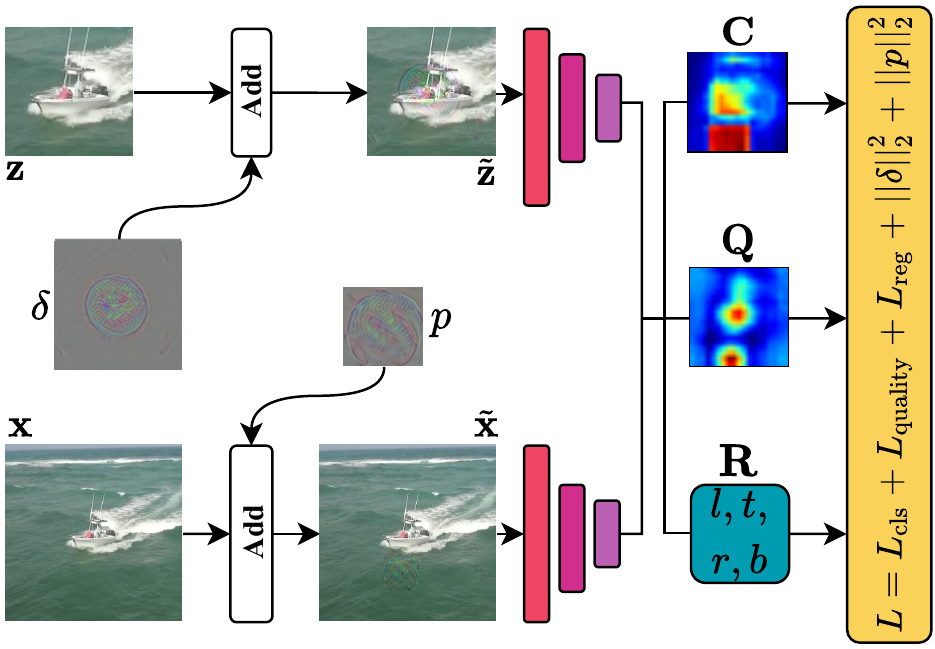}
  \caption{The training pipeline of the proposed method. We aim to train a translucent perturbation $\delta$ for the template image $\textbf z$, and an adversarial patch $p$ for the search image $\textbf x$. After adding $\delta$ to $\textbf z$ and adding $p$ into $\textbf x$, the tracker outputs the location and size of the \textit{fake target} instead of the real target.}
  \label{fig:net}
  \vspace{-3mm}
\end{figure}

\begin{equation}
\begin{gathered}
  \min\limits_{\delta_{\textbf{x}_{i}}, \delta_{\textbf{z}_{i}}} L(y^{fake}_i, f(\textbf{x}_i + \delta_{\textbf{x}_{i}}, \textbf{z}_i + \delta_{\textbf{z}_{i}}))\\
  \text{s.t.}\ ||\delta_{\textbf{x}_i}||_{\mathsf{p}} \le \epsilon_1, ||\delta_{\textbf{z}_i}||_{\mathsf{p}} \le \epsilon_2,
\end{gathered}
\end{equation}
where $L$ is the loss function of the SiamFC++ tracker. $(\textbf{x}_i, \textbf{z}_i)$ is the search-template pair of the video being tracked at frame $i$ and $\textbf{z}_i \equiv \textbf{z}_1$. $y^{fake}_i=\{\textbf{C}^*, \textbf{R}^*, \textbf{Q}^*\}$ is derived from the \textit{fake target} for the loss function and generated according to the fake \uline{ground-truth} box $b^{fake}_i$ that we want the tracker to output at frame $i$ (see Sec. \ref{generate}). Additionally, the perturbation $\delta$ should be sufficiently small, which is commonly modeled through an upper bound $\epsilon$ on the $l_{\mathsf{p}}\text{-norm}$, commonly denoted as $||\cdot||_{\mathsf{p}}$. A popular choice is to set ${\mathsf{p}}=\infty$. However, these methods require perturbing the input image pair $(\textbf{x}, \textbf{z})$ frame by frame, which comes at a non-negligible cost, considering that visual object tracking is a real-time task. As a consequence, universal adversarial perturbations (UAPs) \cite{UAP, shafahi2020universal} are more practical for attacking Siamese trackers:
\begin{equation}
  \begin{gathered}
    \min\limits_{\delta_\textbf{x}, \delta_\textbf{z}} \mathop{\mathbb{E}}\limits_{(\textbf{x}, \textbf{z})} L(y^{fake}, f(\textbf{x} + \delta_\textbf{x}, \textbf{z} + \delta_\textbf{z}))\\
    \text{s.t.}\ ||\delta_\textbf{x}||_{\mathsf{p}} \le \epsilon_1, ||\delta_\textbf{z}||_{\mathsf{p}} \le \epsilon_2,
  \end{gathered}
  \label{eq:UAP}
\end{equation}  
where the input pair $(\textbf x,\textbf z)$ is randomly picked from the training set. Note that universal adversarial perturbations are added to the whole \uline{images} and $y^{fake}$ is kept constant during the training process. As shown in \cite{hirano2020simple}, this vanilla UAP method is limited to untargeted attacks. To achieve universal targeted attacks, one feasible way is to paste a small (universal) adversarial patch into the search images adhering to the predefined trajectory, so that the tracker outputs the location and size of the adversarial patch instead of the real target:
\begin{equation}
    \min\limits_{p_\textbf{x}} \mathop{\mathbb{E}}\limits_{(\textbf{x}, \textbf{z}, y^{fake})} L(y^{fake}, f(A_{\text{paste}}(\textbf{x}, p_\textbf{x}, b^{fake}_{\textbf{x}}), \textbf{z})),
\end{equation}
where $A_{\text{paste}}$ is a patch application operator \cite{patch} which pastes the adversarial patch $p_\textbf{x}$ into the search image according to $b^{fake}_{\textbf{x}}$\uline{, and $b^{fake}_{\textbf{x}}$ denotes the \textit{fake target} in} the search image. $y^{fake}$ is generated according to $b^{fake}_{\textbf{x}}$ and both of them are variables during training. More specifically, $A_\text{paste}$ means that, in the region where the perturbation is pasted, the pixel values of the original image are \textit{replaced} with the pixel values of the perturbation. Note that this vanilla adversarial patch method uses the clean template image during training. However, CNN attacks are usually expected to be imperceptible whereas the above method has to paste an obviously noticeable \textit{fake target} patch into tracking frames, which raises the risk of being suspected.

To overcome the aforementioned shortcomings, we propose to train a translucent perturbation $\delta$ for the template image $\textbf z$, and a translucent patch $p$ for the search image $\textbf x$. After adding $\delta$ to $\textbf z$ and adding the \textit{fake target} patch $p$ into $\textbf x$, the tracker outputs the location and size of the adversarial patch instead of the real target (see Fig.~\ref{fig:net}). Both $\delta$ and $p$ are universal (\ie, video-agnostic), which means perturbing a novel video only involves the mere addition of the perturbations to the template and search images -- and does not require gradient optimization or network inference. This is achieved by
\begin{equation}
  \begin{gathered}
    \min\limits_{p_\textbf{x}, \delta_\textbf{z}} \mathop{\mathbb{E}}\limits_{(\textbf{x}, \textbf{z}, y^{fake})} L(y^{fake}, f(A_{\text{add}}(\textbf{x}, p_\textbf{x}, b^{fake}_{\textbf{x}}), \textbf{z} + \delta_\textbf{z}))\\
    \text{s.t.}\ ||p_\textbf{x}||_{\mathsf{p}} \le \epsilon_1, ||\delta_\textbf{z}||_{\mathsf{p}} \le \epsilon_2,
  \end{gathered}
\end{equation}
where $A_{\text{add}}$ is a patch application operator which adds \uline{a small} patch into the search image according to $b^{fake}_{\textbf{x}}$. \uline{This also departs from Eq.~\eqref{eq:UAP} in that the universal adversarial perturbation $\delta_\textbf{x}$ is added to the whole search image $\textbf{x}$ and limited to untargeted attacks. More specifically, the} operator $A_\text{add}$ refers to the injection of our translucent adversarial patch into the search image, \ie, the values of our translucent adversarial patch are \textit{added} to the pixel values of the original image in the $b^{fake}_{\textbf{x}}$ area.

Our attacks \uline{to} Siamese visual tracking with both the template and search images as input, enable us to exploit the advantages of both the perturbation generation method \cite{FGSM} and adversarial patch generation method \cite{patch} simultaneously. However, applying both of the above methods to trackers is non-trivial. In this paper, we demonstrate that the perturbation and the adversarial patch are both indispensable for attacking trackers and can be jointly optimized in an end-to-end training manner. First, the small patch injected into the search image works as a \textit{fake target}. Since it is crucial to design a strategy for the tracker to follow a predefined trajectory, the TTP method \cite{TTP} achieves this goal by pre-computing perturbations corresponding to diverse directions and using them to force the tracker to follow the predefined trajectory. However, this method is sub-optimal because the attacked tracker only predicts the approximation of the specified trajectory. As shown in our experiments, the targeted attack performance of TTP is not as good as ours. Second, the adversarial patch is usually added to the background region of the search image and does not change the pixel values in the foreground region where the real target exists. Therefore if we do not add additional perturbation to the template image, the response values \uline{in} the region where the real target exists on the heatmaps are always high. Thus it is necessary to perturb the template image to \uline{cool} down hot regions where the real target exists and \uline{increase} the responses at the position of the \textit{fake target}.
\vspace{-3mm}

\begin{figure}[t]
  \centering
  \includegraphics[width=0.45\textwidth]{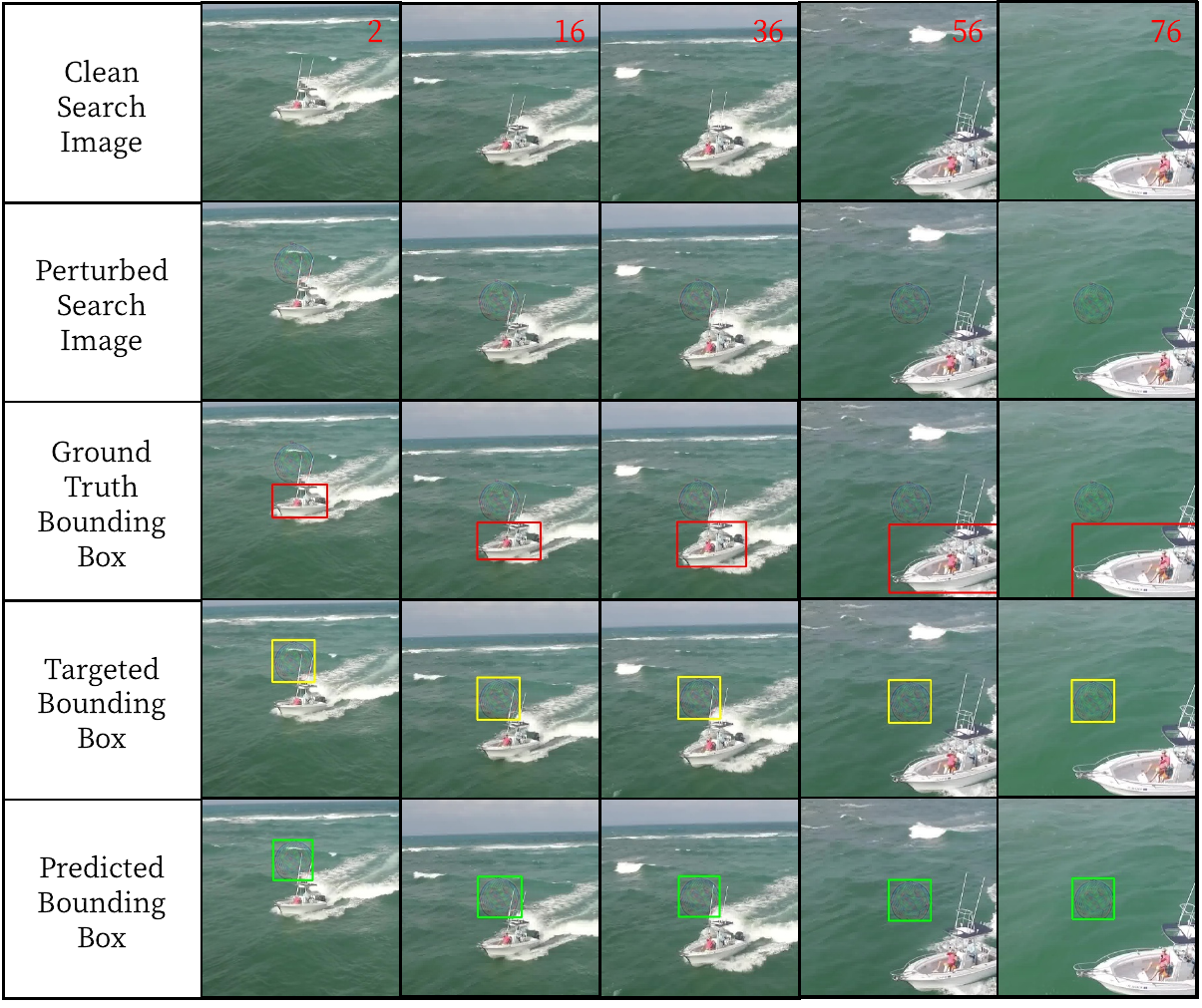}
  \caption{Results of targeted attacks where the tracker is forced to follow a predefined \textit{fake trajectory} $B^{fake}=\{b^{fake}_i\}_1^{T}$ (indicated by the yellow bounding boxes). The \textit{fake trajectory} follows the real trajectory $B^{gt}=\{b^{gt}_i\}_1^T$ (indicated by the red bounding boxes) except that the adjacent boundaries of $b^{fake}_i$ and $b^{gt}_i$ are 2 pixels apart.}
  \label{fig:vis1}
  \vspace{-3mm}
\end{figure}

\subsection{Generating Video-Agnostic Perturbations}\label{generate}

In this subsection, we show how to train our video-agnostic perturbations $(\delta, p)$ for Siamese trackers. At beginning, each element in $\delta$ and $p$ is initialized to 0. During the $k$-th iteration of training, a video $V=\{I_i\}_1^T$ is randomly selected from the training dataset $\mathcal V$. Assuming the template perturbation at the $k$-th iteration is $\delta_k \in \mathbb{R}^{127\times 127 \times 3}$, and the adversarial patch is $p_k$. We first randomly pick paired frames $I_t, I_s$ from $V$. The clean template image $\textbf z\in \mathbb{R}^{127\times 127 \times 3}$ is generated according to $I_t$ and $b^{gt}_t$, and the perturbed template image is:
\begin{equation}
\tilde {\textbf z} = \textbf z + \delta_k.
\end{equation}
Similarly, the clean search image $\textbf x \in \mathbb{R}^{303\times 303 \times 3}$ is generated according to $I_s$ and $b^{gt}_s$. As mentioned before, the patch is regarded as a \textit{fake target} and added into the search images. We force the center position of the \textit{fake target} to near the center position of the real target within a shift range of 64 pixels, where the shift is defined as the maximum range of translation generated from a uniform distribution. The perturbed search image is generated as follows:
\begin{equation}
\tilde{\textbf x} = A_{\text{add}}(\textbf x, p_k, b^{fake}_k),
\end{equation}
where $ b^{fake}_k = \{x_0, y_0, x_1, y_1\}$ denotes the coordinates of the upper-left and lower-right corners of the \textit{fake target} in the search image, \uline{and} $A_\text{add}$ adds the patch into $\textbf x$ at location $(\frac{x_0+x_1}{2},\frac{y_0+y_1}{2})$. Subsequently, the SiamFC++ tracker takes $\tilde {\textbf x}$ and $\tilde{\textbf z}$ as input and predicts $\textbf{C, R, Q}$ in an anchor-free manner.

\textit{Generating Fake Labels.} The fake labels are composed of three parts: fake classification label $\textbf{C}^*$, fake regression label $\textbf{R}^*$ and fake quality estimation label $\textbf{Q}^*$.

For classification, a location $(x,y)$ in the feature map $\psi_{\mathrm{cls}}$ is considered as a positive sample (\ie, $\textbf{C}^*_{x,y} = 1$) if its corresponding location $\left(\left\lfloor\frac{s}{2}\right\rfloor+x s,\left\lfloor\frac{s}{2}\right\rfloor+y s\right)$ in the input image falls into the fake bounding box, and considered as a negative sample (\ie, $\textbf{C}^*_{x,y} = 0$) vice versa. $s=8$ denotes the total step length of the feature extraction network.

In the regression branch, the last convolution layer predicts the distance of a input image position $\left(\left\lfloor\frac{s}{2}\right\rfloor+x s,\left\lfloor\frac{s}{2}\right\rfloor+y s\right)$ to the four edges of the fake bounding box, denoted as the four-dimensional vector $\boldsymbol{t}^{*}=\left(l^{*}, t^{*}, r^{*}, b^{*}\right)$. Thus, the fake regression label $\textbf{R}^*$ at location $(x,y)$ can be expressed as:

\begin{equation}
  \begin{array}{ll}
  l^{*}=\left(\left\lfloor\frac{s}{2}\right\rfloor+x s\right)-x_{0}, \quad t^{*}=\left(\left\lfloor\frac{s}{2}\right\rfloor+y s\right)-y_{0} \\
  r^{*}=x_{1}-\left(\left\lfloor\frac{s}{2}\right\rfloor+x s\right), \quad b^{*}=y_{1}-\left(\left\lfloor\frac{s}{2}\right\rfloor+y s\right)
  \end{array}
\end{equation}

SiamFC++ assumes that feature locations near the center of the target will have much more importance than other locations. Following the design of SiamFC++, we use a $1 \times 1$ convolution layer for quality assessment, \ie, to learn the intersection over union (IoU) score of the predicted box $b^{pred}_k$ and $b^{fake}_k$. Thus, the fake quality estimation label $\textbf{Q}^*$ at location $(x,y)$ can be expressed as:

\begin{equation}
  \mathrm{IoU}^{*}=\frac{\operatorname{ Intersection }\left(b^{pred}_k, b^{fake}_k\right)}{\operatorname{Union}\left(b^{pred}_k, b^{fake}_k\right)}
\end{equation}

\textit{Training Objective.} The loss function is calculated as follows:
\begin{equation}
\begin{array}{l}
\begin{aligned}
L&=\frac{\alpha}{N_{\mathrm{pos}}} \sum_{x, y} L_{\mathrm{cls}}\left(\textbf{C}_{x, y}, \textbf{C}_{x, y}^{*}\right) \\
&+\frac{\beta}{N_{\mathrm{pos}}} \sum_{x, y} \textbf{1}_{\left\{\textbf{C}_{x, y}^{*}>0\right\}} L_{\mathrm{quality}}\left(\textbf{Q}_{x, y}, \textbf{Q}_{x, y}^{*}\right) \\
&+\frac{\gamma}{N_{\mathrm{pos}}} \sum_{x, y} \textbf{1}_{\left\{\textbf{C}_{x, y}^{*}>0\right\}} L_{\mathrm{reg}}\left(\textbf{R}_{x, y}, \textbf{R}_{x, y}^{*}\right) \\
&+\eta_1 \cdot ||\delta_k||_2^2 +  \eta_2 \cdot ||p_k||^2_2,
\end{aligned}
\end{array}
\label{eq:loss}
\end{equation}
where $\textbf{C}_{x, y}, \textbf{R}_{x, y}, \textbf{Q}_{x, y}$ represent the values of $\textbf{C}, \textbf{R}, \textbf{Q}$ at location $(x, y)$ respectively, and $\textbf{C}^*, \textbf{R}^*, \textbf{Q}^*$ are the fake labels generated according to the position and size of the \textit{fake target}. $\textbf 1$ is the indicator function that takes 1 if the condition in subscribe holds and takes 0 if not. $N_{\mathrm{pos}}$ denotes the number of positive samples in the training phase, $L_{\mathrm{cls}}$ denotes the focal loss \cite{focal} for classification result,
$L_{\mathrm{quality}}$ denotes the binary cross entropy (BCE) loss for quality assessment, and $L_{\mathrm{reg}}$ denotes the IoU loss \cite{iou-loss} for bounding box regression.
 
\begin{algorithm}[tb]
  \small
  \caption{Training Process}
  \label{alg:algorithm}
  \textbf{Input}: Training dataset $\mathcal{V}$, Siamese tracker $f$, and max iteration number $N$.\\
  \textbf{Output}: $\delta$, $p$.
  \begin{algorithmic}[1] 
  \STATE Let $k = 0$.
  \WHILE{$k < N$}
  \STATE Randomly pick a video $V\in \mathcal{V}$. The corresponding \uline{ground-truth} is $B^{gt}=\{b^{gt}_i\}^T_1$.
  \STATE Randomly pick paired frames $I_t, I_s$ from $V$.
  \STATE Generate template image $\textbf{z}$ according to $I_t$ and $b^{gt}_t$.
  \STATE $\tilde{\textbf{z}} = \textbf{z} + \delta_k.$
  \STATE Generate search image $\textbf{x}$ according to $I_s$ and $b^{gt}_s$.
  \STATE Calculate the \textit{fake target} position $\{x_0, y_0, x_1, y_1\}$ with respect to the search image.
  \STATE $\tilde{\textbf x} = A_{\text{add}}(\textbf x, p_k, \{x_0, y_0, x_1, y_1\}).$
  \STATE $\textbf{C, R, Q} = f(\tilde {\textbf x}, \tilde{\textbf z}).$
  \STATE Generate fake labels $\textbf{C}^*,\textbf{R}^*,\textbf{Q}^*$ using $\{x_0, y_0, x_1, y_1\}$.
  \STATE Calculate loss $L(\textbf{C, R, Q}, \textbf{C}^*, \textbf{R}^*, \textbf{Q}^*)$ using Equ. \ref{eq:loss}.
  \STATE $\delta_{k+1} = \delta_{k} - \epsilon_1 \cdot \text{sign}(\nabla_{\delta_k}L).$
  \STATE $p_{k+1} = p_{k} - \epsilon_2 \cdot \text{sign}(\nabla_{p_k}L).$
  \STATE $k = k + 1.$
  \ENDWHILE
  \STATE \textbf{return} $\delta_N, p_N.$
  \end{algorithmic}
  \label{alg}
\end{algorithm}
  
\textit{Optimization.}
Before introducing our perturbation update process in offline training, we first revisit the popular adversarial example generation methods (\eg,~\cite{FGSM, DBLP:conf/iclr/KurakinGB17a}). One of the simplest methods to generate adversarial image $I^{adv}$ is FGSM~\cite{FGSM} and works by linearizing the loss function around the network weights and obtaining an optimal max-norm constrained perturbation for generating the adversarial image:
\begin{equation}
    I^{adv} = I + \epsilon \cdot \text{ sign} \bigl( \nabla_I J(I, y_{true})  \bigr),
    \vspace{-0.1cm}
\end{equation}
where $I$ is the input clean image, and the values of its pixels are integer numbers in the range [0, 255]. $y_{true}$ is the true label for the image $I$, $J(I, y_{true})$ is the cost function for training the neural network, and $\epsilon$ is a hyper-parameter to be chosen. A straightforward way to extend the above method is applying it multiple times with small step size. This leads to Basic Iterative Method (BIM) introduced in \cite{DBLP:conf/iclr/KurakinGB17a}:
\begin{equation}
    \begin{gathered}
        I_0^{adv} = I, \\
        I_{N+1}^{adv} = Clip_{I, \epsilon}\left\{I_N^{adv}+\epsilon \cdot \text{ sign}(\nabla_I J(I_N^{adv},y_{true}))\right\},
    \end{gathered}
\end{equation}
where the pixel values of intermediate results are clipped after each step to ensure that they are in an $\epsilon$-neighbourhood of the original image. The BIM method can be easily made into an attacker for a specific desired target class, called Iterative Target Class Method \cite{DBLP:conf/iclr/KurakinGB17a}:
\begin{equation}
  \begin{gathered}
      I_0^{adv} = I,\\
      I_{N+1}^{adv} = Clip_{I, \epsilon}\left\{I_N^{adv}-\epsilon \cdot \text{ sign}(\nabla_I J(I_N^{adv},y_{target}))\right\}.
  \end{gathered}
  \label{equ:itcm}
\end{equation}

We utilize this Iterative Target Class Method to update our perturbation values during the training process. To achieve a balance between the attack efﬁciency and the perturbation perceptibility, we constrain the perturbation values in the loss function of Eq.~\eqref{eq:loss} instead of using the clip operation. At each training step, our perturbations are updated as follows:
\begin{gather}
\delta_{k+1} = \delta_{k} - \epsilon_1 \cdot \text{sign}(\nabla_{\delta_k}L)\\
p_{k+1} = p_{k} - \epsilon_2 \cdot \text{sign}(\nabla_{p_k}L),
\end{gather}
where $\epsilon_1$ and $\epsilon_2$ are to ensure that the perturbation added to the template/search image is small. During training, we only optimize the values of $(\delta, p)$ and the network weights remain intact. We outline this training procedure in Algorithm \ref{alg}.

\subsection{Attacking the Tracker at Inference Time}

Once the perturbations $(\delta, p)$ are trained, we can use them to perturb the template and search images of any novel video for attacking. Both $\delta$ and $p$ are universal (\ie, video-agnostic), which means perturbing a novel video only involves the mere addition of the perturbations to the template and search images -- and does not require gradient optimization or network inference. Assume $B^{fake}=\{b^{fake}_i\}_1^{T}$ is the trajectory we hope the tracker to output. During tracking the $i$-th frame of the video $V=\{I_i\}_1^T$, we need to add $p$ into $\textbf x_i$ according to $b^{fake}_i=\{x_{0_i}, y_{0_i}, x_{1_i}, y_{1_i}\}$:
\begin{equation}
\tilde{\textbf x}_i = A_{\text{add}}(\textbf x_i, p, \{x_{0_i}, y_{0_i}, x_{1_i}, y_{1_i}\}).
\end{equation}
The tracker then takes $\tilde{\textbf z}_1=\textbf z_1+\delta$ and $\tilde{ \textbf x}_i$ as input, and the subsequent tracking procedure remains the same as SiamFC++. We outline this procedure in Algorithm \ref{alg:algorithm_attack}.

\begin{algorithm}[tb]
  \small
  \caption{Attack Process}
  \label{alg:algorithm_attack}
  \textbf{Input}: The trained perturbations $\delta$ and $p$, Siamese tracker $f$, video $V=\{I_i\}_1^T$. $b^{gt}_1$ is the position of the real target in the first frame. $B^{fake}=\{b^{fake}_i\}_1^{T}$ is the trajectory we hope the tracker to output.\\
  \textbf{Output}: $B^{pred}=\{b^{pred}_i\}_1^{T}$
  \begin{algorithmic}[1] 
    \STATE Generate the clean template image $\textbf{z}_1$ according to $I_1$ and $b^{gt}_1$.
    \STATE Generate the perturbed template image $\tilde{\textbf z}_1=\textbf z_1+\delta$.
    \STATE Let $i = 2$.
  \WHILE{$i \le T$}
  \STATE Generate clean search image $\textbf{x}_i$ according to $I_i$ and $b^{pred}_{i-1}$.
  \STATE $b^{fake}_i=\{x_{0_i}, y_{0_i}, x_{1_i}, y_{1_i}\}$
  \STATE Generate the perturbed search image $\tilde{\textbf x}_i = A_{\text{add}}(\textbf x_i, p, \{x_{0_i}, y_{0_i}, x_{1_i}, y_{1_i}\}).$
  \STATE $\textbf{C, R, Q} = f(\tilde {\textbf x}_i, \tilde{\textbf z}_1).$
  \STATE Generate the predicted bounding box $b^{pred}_i$ according to $\textbf{C, R, Q}$.
  \STATE $i = i + 1.$
  \ENDWHILE
  \STATE \textbf{return} $B^{pred}$
  \end{algorithmic}
\end{algorithm}

\section{Experiments}
\subsection{Experimental Setup}\label{setup}
\begin{table*}[t]
  \centering
  \caption{Characteristics of the datasets used to train and evaluate the proposed attack method.}
  \begin{tabular}{lrccccc} \toprule
  \multicolumn{2}{c}{Dataset}                            & Videos & Total frames & Frame rate & Object classes & Num. of attributes \\ \midrule
  \multirow{4}{*}{Training set} & GOT-10k training split & 9.34K  & 1.4M        & 10 fps     & 480            & 6                  \\
                                & LaSOT training split   & 1.12K  & 2.83M        & 30 fps     & 70             & 14                 \\
                                & COCO2017               & n/a    & 118K         & n/a        & 80             & n/a                \\
                                & ILSVRC-VID             & 5.4K   & 1.6M         & 30 fps     & 30             & n/a                \\ \midrule
  \multirow{7}{*}{Test set}     & GOT-10k validation split& 180   & 21K          & 10 fps     & 150            & 6                  \\
                                & LaSOT test split       & 280    & 690K         & 30 fps     & 70             & 14                 \\
                                & OTB-15                 & 100    & 59K          & 30 fps     & 22             & 11                 \\
                                & VOT2016                & 60     & 21K          & 30 fps     & 16             & 6                  \\
                                & VOT2018                & 60     & 21K          & 30 fps     & 24             & 6                  \\ 
                                & VOT2019                & 60     & 19K          & 30 fps     & 30             & 6                  \\ \bottomrule
  \end{tabular}
  \vspace{-4mm}
  \label{tab:dataset}
\end{table*}
\textit{Evaluation Benchmarks.} We evaluate our video-agnostic perturbation method for targeted attacks on several tracking benchmarks, \ie, OTB2015 \cite{OTB}, GOT-10k \cite{GOT-10k}, LaSOT \cite{LaSOT}, VOT2016 \cite{VOT2016}, VOT2018 \cite{VOT2018} and VOT2019 \cite{VOT2019}.
Generally speaking, OTB2015 is a typical tracking benchmark which is widely used for evaluation for several years. GOT-10k has the advantage of magnitudes wider coverage of object classes. LaSOT has much longer video sequences with an average duration of 84 seconds. OTB2015, GOT-10k and LaSOT all follow the One-Pass Evaluation (OPE) protocol and their evaluation methodologies are similar as the measurement is mostly based on the success and precision of the trackers over the test videos. For instance, they all measure the success based on the fraction of frames in a sequence where the intersection-over-union (IoU) overlap of the predicted and \uline{ground-truth} rectangles exceeds a given threshold, and then the trackers are ranked using the area-under-the-curve (AUC) criterion. Since the average of IoU overlaps (AO) over all the test video frames is recently proved to be equivalent to the AUC criterion, we thus denote the success measurement as AO in the following. Besides AO, a success rate (SR) metric is also directly used to measure the percentage of successfully tracked frames given a threshold as in GOT-10k. As for the precision, it encodes the proportion of frames for which the center of the predicted rectangle is within 20 pixels of the \uline{ground-truth} center. Since the precision metric is sensitive to the resolution of the images and the size of the bounding boxes, a metric of normalized precision over the size of the \uline{ground-truth} bounding box is proposed and the trackers are then ranked using the AUC for normalized precision between 0 and 0.5. VOT \cite{VOT2016,VOT2018,VOT2019} introduces a series of tracking competitions with up to 60 sequences in each of them, aiming to evaluate the performance of a tracker in a relatively short duration. Different from other datasets, the VOT dataset has a reinitialization module. When the tracker loses the target (\ie, the overlap is zero between the predicted result and the annotation), the tracker will be reinitialized for the remaining frames based on the \uline{ground-truth} annotation. Three metrics are used to evaluate the performance of a tracker in VOT: (1) accuracy, (2) robustness and (3) EAO (expected average overlap). The accuracy measures how well the bounding box predicted by the tracker overlaps with the \uline{ground-truth} bounding box. The robustness measures how many times the tracker loses the target (fails) during tracking. EAO combines accuracy and robustness to evaluate the overall performance of the tracker. Characteristics of these datasets are summarized in Table \ref{tab:dataset}.

\textit{Generating the Fake Trajectory.} We need to predefine a specific trajectory $B^{fake}=\{b^{fake}_i\}_1^{T}$ for each video to achieve targeted attack in the online-tracking phase, which we call the \textit{fake trajectory}. We denote the real trajectory as $B^{gt}=\{b^{gt}_i\}_1^T$ with the \uline{ground-truth} bounding boxes. It is possible to manually label arbitrary $B^{fake}$ for each video, however, it will be time-consuming in our experimental evaluation. So we generate $B^{fake}$ based on $B^{gt}$. Specifically, the \textit{fake trajectory} follows the real trajectory except that the adjacent boundaries of $b^{fake}_i$ and $b^{gt}_i$ are 2 pixels apart (Fig. \ref{fig:vis1}). Because the annotations of GOT-10k's test split are kept private, we choose to use its validation set for our targeted attack evaluation and denote it as GOT-Val. Note that the target position of the first frame is \uline{predefined} in the academic study of object tracking, while in practical \uline{applications}, the target position of the first frame is often obtained by running an object detector on this frame. In this scenario, once the target position of the first frame is obtained, we can put our adversarial patch near the target to mislead the tracker. In the remaining frames, we can specify an arbitrary \textit{fake trajectory} to place the adversarial patch.

\textit{Image Quality Assessment.} We use structural similarity (SSIM) \cite{SSIM} to evaluate the quality and perceptibility of our generated perturbations $\delta$ and $p$. It is difficult to perceive the perturbation when SSIM is close to 1 (see Table \ref{tab:iter}).

\subsection{Implementation Details}

In our evaluation, the backbone Siamese network of our base tracker SiamFC++ \cite{SiamFC++} adopts GoogLeNet \cite{GoogLeNet}. We implement our approach in Pytorch and train our perturbations using three GTX 1080Ti GPUs. We adopt COCO \cite{COCO}, ILSVRC-VID \cite{VID} and the training splits of GOT-10k \cite{GOT-10k} and LaSOT \cite{LaSOT} as our training set. We train the perturbations for 8192 iterations with a mini-batch of 96 images (32 images per GPU). Both the hyper-parameters $\epsilon_1$ and $\epsilon_2$ for the template perturbation and the adversarial patch are set to 0.1. We generate training samples following the practice in SiamFC++. During both the training and online-tracking \uline{phases}, the off-the-shelf SiamFC++ tracking network model\footnote{It is exclusively trained on the training split of GOT-10k by the authors of SiamFC++ and can be downloaded from \url{https://drive.google.com/file/d/1BevcIEZr_kgyFjhxayOFw08DFl2u5Zi7/view}} is fixed and used for the whole evaluation, the spatial size of the template image is set to $127\times 127$, and the search image is $303\times 303$. In Eq.~\eqref{eq:loss}, we set $\alpha=1, \beta=1, \gamma=1, \eta_1=0.005$, and $\eta_2=0.005$.

\begin{table}[t]
  \centering
  \caption{Overall attack results on VOT2016, VOT2018, VOT2019.}
  \begin{tabular}{rrcc}
  \toprule
  Benchmarks & Metrics & Before Attack    & Untargeted Attack  \\
  \midrule
  \multirow{2}{*}[-6pt]{VOT2016} 
  & Accuracy   & 0.626 & 0.393\\
  & Robustness & 0.144 & 9.061\\
  & EAO        & 0.460 & 0.007\\
  \midrule
  \multirow{2}{*}[-6pt]{VOT2018} 
  & Accuracy   & 0.587 & 0.342\\
  & Robustness & 0.183 & 8.981\\
  & EAO        & 0.426 & 0.007\\
  \midrule
  \multirow{2}{*}[-6pt]{VOT2019} 
  & Accuracy   & 0.556 & 0.345\\
  & Robustness & 0.537 & 8.824\\
  & EAO        & 0.243 & 0.010\\
  \bottomrule
  \end{tabular}
  \label{tab:benchmark results1}
\end{table}

\begin{table}[t]
  \centering
  \caption{Overall attack results on OTB2015, GOT-Val and LaSOT.}
  \begin{tabular}{rrccc}
  \toprule
  \multirow{2}{*}{Benchmarks} & \multirow{2}{*}{Metrics} & Before    & Untargeted & Targeted  \\
                            &                         & Attack & Attack & Attack     \\ 
  \midrule
  \multirow{2}{*}{OTB2015} 
  & AO   & 0.642 & 0.063 & 0.759\\
  & Precision & 0.861 & 0.092 & 0.795\\
  \midrule
  \multirow{2}{*}{GOT-Val} 
  & SR & 0.897 & 0.123 & 0.890\\
  & AO & 0.760 & 0.153 & 0.840 \\
  \midrule
  \multirow{3}{*}{LaSOT} 
  & Precision  & 0.514 & 0.046 & 0.605\\
  & Norm. Prec.& 0.551 & 0.048 & 0.702\\
  & AO         & 0.525 & 0.069 & 0.691\\
  \midrule
  \multicolumn{2}{r}{FPS} & 58 & 58 & 58\\
  \bottomrule
  \end{tabular}
  \label{tab:benchmark results}
  \vspace{-3mm}
\end{table}
\begin{table}[t]
  \centering
  \caption{Comparison of attack performance with 2 baseline methods on GOT-Val in terms of AO. Baseline-1 performs untargeted attacks based on the UAP \cite{UAP} method. Baseline-2 performs targeted attacks based on the adversarial patch \cite{patch} method.}
  \begin{tabular}{@{}rcc@{}}
  \toprule
  Methods & Untargeted Attack & Targeted Attack \\ \midrule
  Baseline-1 \cite{UAP}  & 0.09          & -\\
  Baseline-2 \cite{patch}   & 0.23       & 0.78\\
  Ours & 0.15       & 0.84\\ \bottomrule
  \end{tabular}
  \label{tab:imperceptible}
\end{table}
\begin{figure}[t]
  \centering
  \subfigure[Baseline-1]{\label{fig:imperceptibleb}\includegraphics[width=0.15\textwidth]{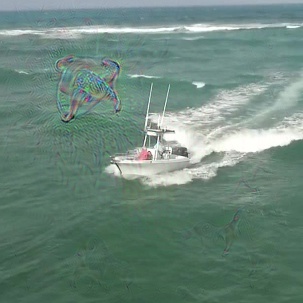}}
  \subfigure[Baseline-2]{\label{fig:imperceptiblec}\includegraphics[width=0.15\textwidth]{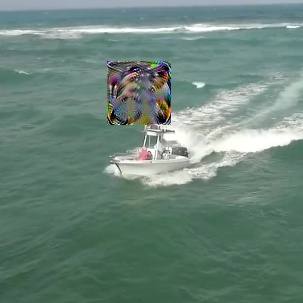}}
  \subfigure[Ours]{\label{fig:imperceptibled}\includegraphics[width=0.15\textwidth]{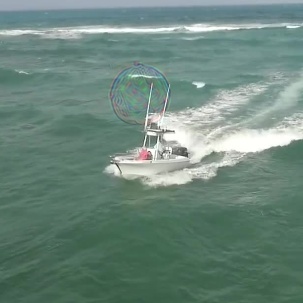}}
  \caption{Visualization of perturbations on search images. Baseline-1 generates \uline{one perturbation} added to the whole search \uline{images} and cannot perform targeted attacks. Baseline-2 pastes \uline{the} obviously noticeable patch into the search \uline{images}, which obviously raises the risk of being suspected. Our method adds a \textit{small} universal patch to the search \uline{images} to perform targeted attacks, which \uline{is} as translucent as in Baseline-1.}
  \label{fig:imperceptible}
  \vspace{-3mm}
\end{figure}
\begin{table*}
  \centering
  \caption{Influence of the number of training iterations on GOT-Val.}
  \begin{tabular}{rcccccccccccccccc} 
  \toprule
  \multicolumn{2}{r}{Iterations}     & 1     & 2     & 4     & 8     & 16    & 32    & 64    & 128   & 256   & 512   & 1024  & 2048  & 4096  & 8192  \\ 
  \midrule
  \multirow{2}{*}{Targeted Attack} & AO    &  0.14 & 0.14 & 0.14 & 0.14 & 0.14 & 0.14 & 0.15 & 0.18 & 0.47 & 0.73 & 0.78 & 0.82 & 0.84 & 0.84  \\
                              & SR    &  0.1 & 0.1 & 0.1 & 0.1 & 0.1 & 0.1 & 0.11 & 0.15 & 0.49 & 0.78 & 0.84 & 0.88 & 0.89 & 0.89    \\ 
  \midrule
  \multirow{2}{*}{Untargeted Attack} & AO   & 0.76 & 0.77 & 0.76 & 0.76 & 0.76 & 0.76 & 0.75 & 0.73 & 0.48 & 0.27 & 0.22 & 0.17 & 0.15 & 0.15    \\
                              & SR   & 0.89 & 0.9 & 0.89 & 0.89 & 0.9 & 0.89 & 0.88 & 0.86 & 0.53 & 0.25 & 0.18 & 0.14 & 0.12 & 0.12    \\ 
  \midrule
  \multicolumn{2}{r}{SSIM of $\delta$}&   1 & 1 & 1 & 1 & 0.99 & 0.99 & 0.97 & 0.94 & 0.88 & 0.84 & 0.82 & 0.81 & 0.8 & 0.79\\
  \midrule
  \multicolumn{2}{r}{SSIM of $p$}      &  0.98 & 0.98 & 0.98 & 0.98 & 0.98 & 0.98 & 0.93 & 0.78 & 0.56 & 0.50 & 0.51 & 0.52 & 0.53 & 0.56\\
  \bottomrule
  \end{tabular}
  \label{tab:iter}
  \vspace{-3mm}
\end{table*}
\begin{figure}[t]
  \centering
  \subfigure[FAN]{\label{fig:a1}\includegraphics[width=0.45\textwidth]{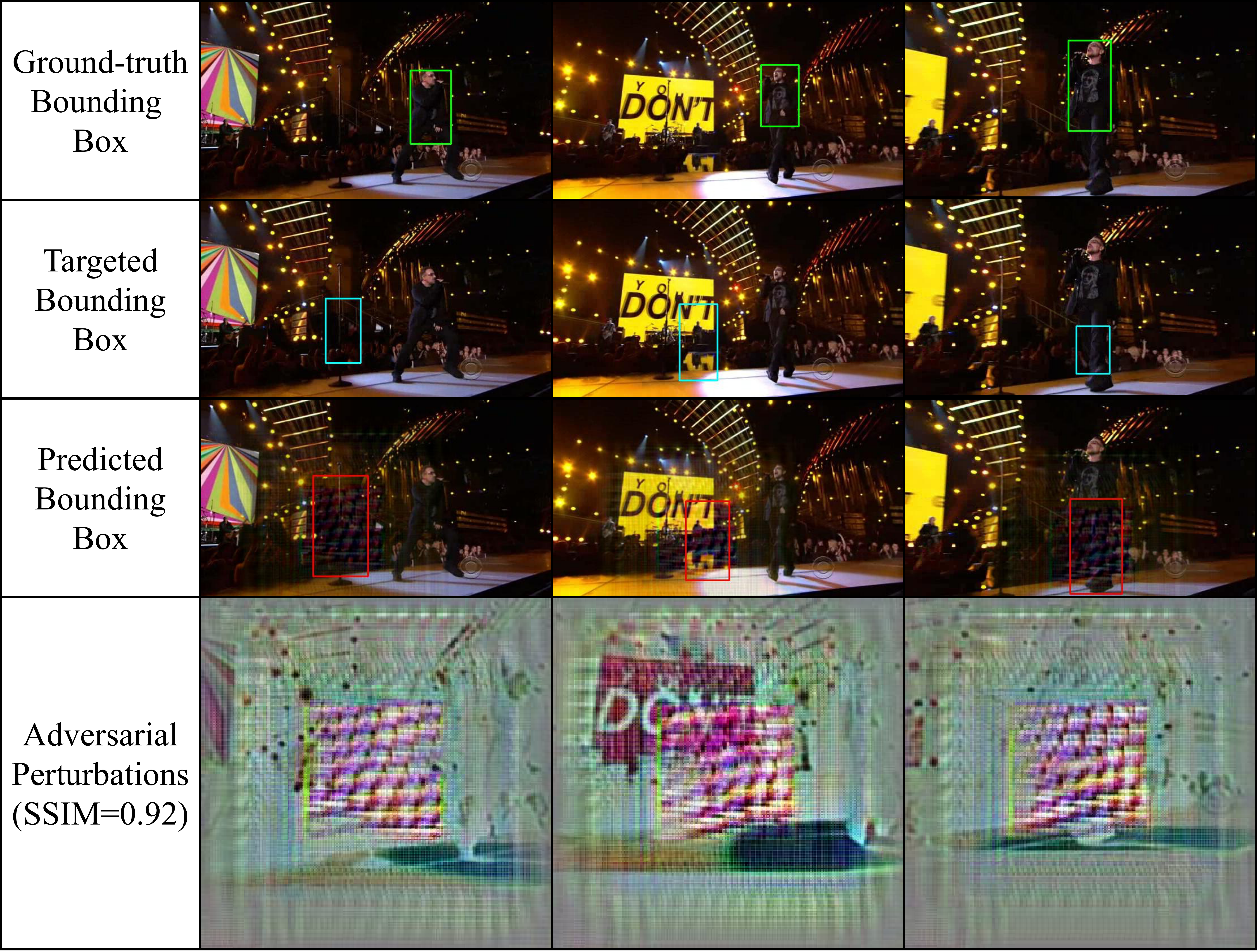}}
  \subfigure[Ours]{\label{fig:b1}\includegraphics[width=0.45\textwidth]{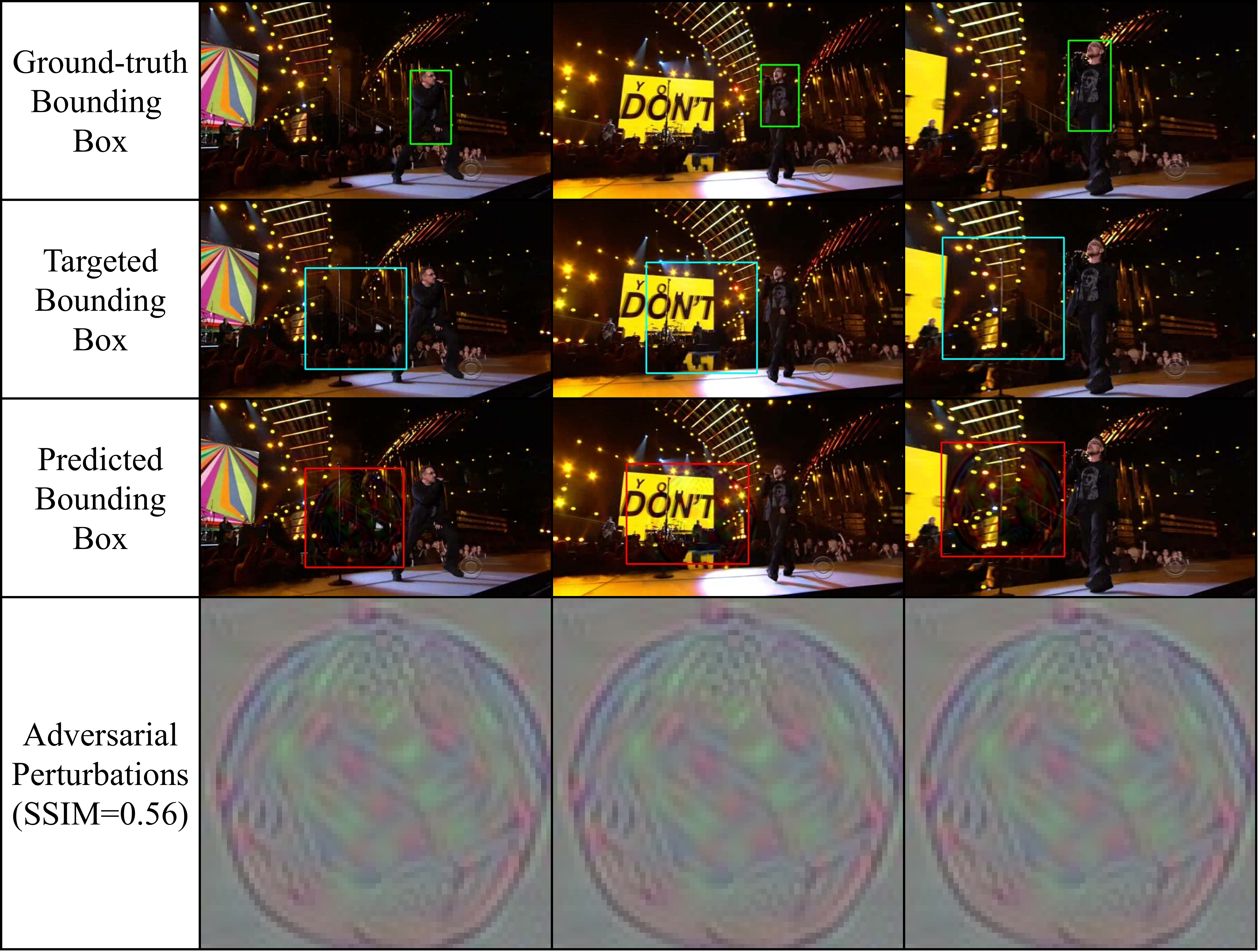}}
  \vspace{-1mm}
  \caption{The results under targeted attacks compared with FAN \cite{FAN}.  FAN crafts the perturbations for each video independently, which comes at a non-negligible computational cost. To achieve video-agnostic universal attacks, we relax the constraint on the value of the perturbations, achieving a balance between the attack efficiency and the perturbation perceptibility.}
  \label{fig:vis_fan}
  \vspace{-4mm}
\end{figure}
\begin{figure}[t!]
  \begin{center}
    \includegraphics[width=0.45\textwidth]{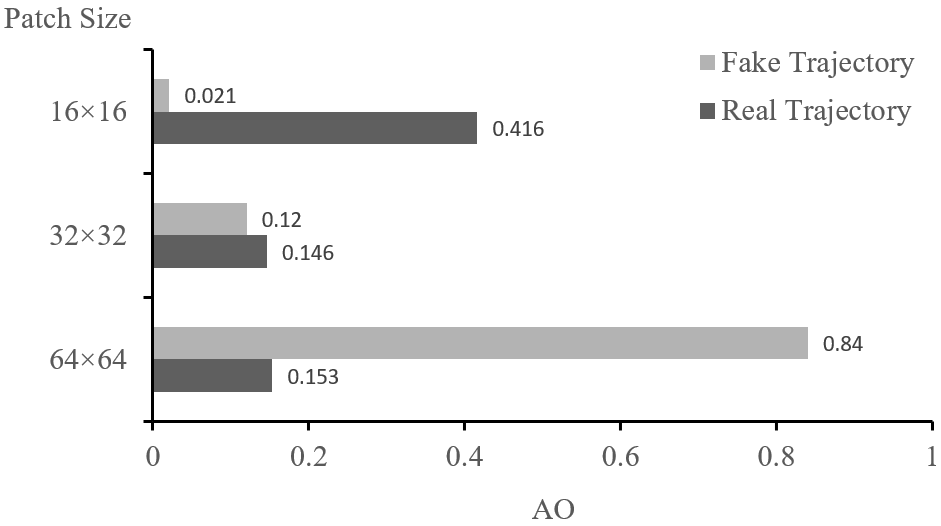}
  \end{center}
  \caption{AO with respect to the real/fake trajectory on the perturbed GOT-Val dataset using three different patch sizes. The \uline{32$\times$32} patch can make the AO score decrease to 0.146 dramatically.}
  \label{fig:patch_size_table}
\end{figure}

\subsection{Overall Attack Results on the Evaluation Benchmarks}

We test the performance of our targeted attack method on the evaluation benchmarks and gather the overall results in Table \ref{tab:benchmark results1} and Table \ref{tab:benchmark results}. \uline{We evaluate the untargeted and targeted attack performance using the following metrics on specific datasets: (1) Accuracy, Robustness and EAO on VOT, (2) AO and Precision on OTB2015, (3) SR and AO on GOT-Val, and (4) Precision, Normalized Precision and AO on LaSOT.} It is shown that the base tracker SiamFC++ can achieve state-of-the-art performance on all the evaluation benchmarks and run in real time (at about 58 fps on a GTX 1080Ti GPU). However this real-time performance requires the computationally intensive tracking system to occupy most of the computational resources, and thus it is appealing to develop a virtually costless attack method to fool the tracking system without scrambling for the resources. As shown in Tables \ref{tab:benchmark results1} and \ref{tab:benchmark results}, our method can satisfy this appealing demand and fool the SiamFC++ tracker effectively by misleading the tracker to follow a predefined \textit{fake trajectory}. Moreover, the high AO and Precision performance calculated by aligning with the \textit{fake trajectory} indicates a more effective targeted attack without raising possible suspicion (see last column of Table \ref{tab:benchmark results}).

\subsection{Analyses of the Perceptibility of Our Perturbations}
We firstly compare our attack method with two baseline methods to show the motivation behind our proposed approach. As introduced in Sec.~\ref{problemdefinition}, our goal is deteriorating the tracking performance of Siamese trackers at a low computational cost as well as making the perturbations less obvious.
Baseline-1 performs untargeted attacks based on the UAP \cite{UAP} method. The difference between Baseline-1 and our method is that, Baseline-1 generates \uline{one perturbation} added to the whole search \uline{images} and cannot perform targeted attacks, while our method adds a \textit{small} universal patch to the search \uline{images} to perform targeted attacks (see Fig. \ref{fig:imperceptible}). Baseline-2 performs targeted attacks based on the adversarial patch \cite{patch} method. The differences between Baseline-2 and our method include: (1) Baseline-2 pastes \uline{the} adversarial patch into the search \uline{images} without the $l_{\mathsf{p}}\text{-norm}$ constraint as ours; (2) Baseline-2 uses the clean template image while we add translucent perturbation to the template image. As a consequence, Baseline-2 generates an obviously noticeable patch while ours is as translucent as in Baseline-1 (see Fig. \ref{fig:imperceptible}). Moreover, our attack performance is also superior to Baseline-2 as shown in Table~\ref{tab:imperceptible}\uline{.}

We also examine the influence of the number of training iterations on the perceptibility of our perturbations. We use SSIM value to indicate the perceptibility, which ranges from 0 to 1. If the SSIM value is close to 1, the difference between the perturbed image and the original image is small, which means it is imperceptible. It can be observed in Table \ref{tab:iter} that as the number of iterations increases, the AO score with respect to the \textit{fake trajectory} increases significantly, while it decreases significantly with respect to the real trajectory. After about \uline{8000} training iterations, the resulting perturbations prevent the tracker from tracking most of the targets in GOT-Val, and the AO with respect to the real trajectory decreases from 0.760 to 0.153. Note that the AO decreases significantly faster at the beginning (training iterations less than 2048). This demonstrates the fast convergence capability of our method. However, the SSIM values of our perturbed images gradually decrease as the training proceeds. At the iteration number of 8192, the SSIM value of our perturbed template image decreases to 0.79, and 0.56 for the perturbed search image. \textit{Note that the SSIM for our perturbed search image is calculated in the sub-region where the patch is placed}.

We further compare our method with FAN \cite{FAN} to examine the perceptibility. FAN \uline{independently generates different perturbations} for each frame, while our perturbations are universal. FAN calculates the average SSIM value of all search images for each dataset, and \textit{the SSIM for their perturbed search image is calculated in the whole area of the search image}. As shown in Fig. \ref{fig:vis_fan}, both FAN and our method can perform targeted \uline{attacks}. Although FAN has its average SSIM value for OTB2015 dataset to only decrease to 0.92, its \uline{perturbations are} also a little noticeable for some video sequences as shown in Fig. \ref{fig:a1}. Note that Fig. \ref{fig:a1} is directly borrowed from their paper as the original perturbation results are not available. Moreover, our universal perturbations achieve better targeted attack performance than FAN on OTB2015 \uline{as shown in Sec.~\ref{SOTA}}.

We note that it is important to make the size of the adversarial patch get small to avoid possible suspicion. So we experimentally examine the influence of the adversarial patch size in Fig. \ref{fig:patch_size_table}. We compare three different sizes of the adversarial patch, \ie, 64$\times$64, 32$\times$32 and 16$\times$16, to conduct this experiment. We evaluate the AO score of SiamFC++ on the perturbed GOT-Val dataset using these three different patch sizes.
As shown in \uline{Fig.} \ref{fig:patch_size_table}, the 32$\times$32 patch can make the AO score decrease to 0.146 dramatically and its area only accounts for 1\% of the search image area.

Since we can only achieve a balance between the attack efficiency and the perturbation perceptibility, our universal perturbation method may be a double-edged sword as it may result in suspicious attacks. Prior works commonly train a network to prevent not only suspicious attacks but also modifying every pixel. So we also examine a new strategy to perturb the search \uline{images} so as to reduce the possible suspicion. Specifically, we first convert \uline{one} search image into YCbCr color space, and then add a perturbation to the entire search image in the Y channel and a different perturbation to a very small region of $64\times64$ in the CbCr \uline{channels}. Different from RGB color space, YCbCr color space encodes a color image similar to human eyes’ retina, which separates the RGB components into a luminance component (Y) and two chrominance components (Cb as blue projection and Cr as red projection). We choose YCbCr color space since its color
channels~~are~~less~~correlated~~than~~RGB~~\cite{8630918}.~~Moreover,~~the
\begin{figure}[t]
  \centering
  \includegraphics[width=0.48\textwidth]{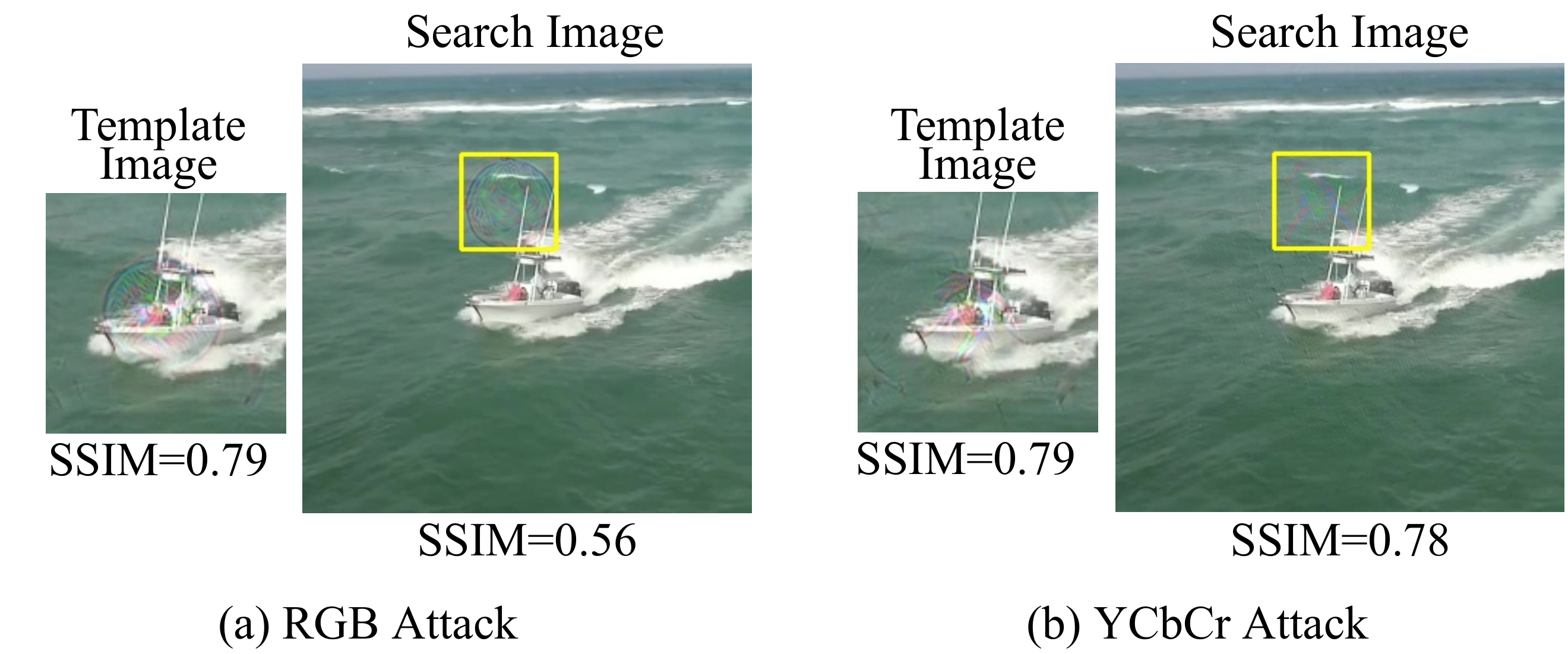}
  \caption{Visualization of the perturbations in attacking the different RGB and YCbCr color spaces. Note that the perturbation \uline{for} the search \uline{images} in the Y channel results in SSIM value of 0.95 outside the added patch area, which \uline{indicates good} imperceptibility.}
  \label{fig:YCbCr}
\end{figure}
\begin{table}[t]
  \centering
  \caption{Comparison between attacking the different RGB and YCbCr color spaces. Attacking the YCbCr color space means we first convert \uline{one} search image into YCbCr color space, and then add a perturbation to the entire search image in the Y channel and a different perturbation to a very small region of $64 \times 64$ in the CbCr \uline{channels}. The attack performance is evaluated on GOT-Val.}
  \label{table:perturb}
  \begin{tabular}{@{}rcccccc@{}}
  \toprule
  \multirow{2}{*}[-1pt]{\begin{tabular}[c]{@{}c@{}}Different Color\\ Space Attack\end{tabular}} & \multicolumn{2}{c}{Untargeted Attack} & \multicolumn{2}{c}{Targeted Attack} & \multicolumn{2}{c}{SSIM} \\ \cmidrule{2-7}
                                                         & AO                                      & SR                               & AO                & SR                   & $\delta$          & $p$  \\ \midrule
  RGB Attack                                             & 0.153                                   & 0.123                            & 0.840             & 0.890                & 0.56              & 0.79 \\
  YCbCr Attack                                           & 0.246                                   & 0.227                            & 0.682             & 0.756                & 0.78              & 0.79 \\ \bottomrule        
  \end{tabular}
  \vspace{-10mm}
\end{table}
\begin{table}[!ht]
  \centering
  \caption{Attack performance comparison using other random patterns. It is evaluated on GOT-Val.}
  \label{table:noise}
  \begin{tabular}{@{}rcccc@{}}
  \toprule
  \multirow{2}{*}[-1pt]{\begin{tabular}[c]{@{}c@{}}Perturbations used to \\ perform attack \end{tabular}} & \multicolumn{2}{c}{Untargeted Attack} & \multicolumn{2}{c}{Targeted Attack}\\ \cmidrule{2-5}
                                                         & AO                                      & SR                               & AO                & SR                  \\ \midrule
  Trained Perturbations                                  & 0.153                                   & 0.123                            & 0.840             & 0.890               \\
  Similar Pattern                                         & 0.736                                   & 0.871                            & 0.153             & 0.118               \\
  Gaussian Noise                                         & 0.740                                   & 0.875                            & 0.144             & 0.101               \\ \bottomrule
  \end{tabular}
\end{table}

\noindent
CbCr channels are less sensitive to the human vision system than the Y channel \cite{8630918}, which means that the patch added in the CbCr channels has better performance of transparency. For the template image, we also convert it into YCbCr color space, and then add the perturbation to the entire template image in all YCbCr channels. The perturbed template and search images are finally converted into RGB color space and fed into the tracking network. The other steps of the training process are the same to the attack method in Sec.~\ref{method}. As shown in Table \ref{table:perturb}, the performance of attacking the YCbCr color space is slightly degraded compared to the RGB color space. However, attacking the YCbCr color space brings better imperceptibility (see Fig. \ref{fig:YCbCr}). Compared with the RGB space, the SSIM value for the search \uline{images} increases from 0.56 to 0.78. Note that the perturbation \uline{for} the search \uline{images} in the Y channel results in SSIM value of 0.95 outside the added patch area, which \uline{indicates good} imperceptibility.
\subsection{Other Analyses}
\begin{figure}[t]
  \centering
  \includegraphics[width=0.48\textwidth]{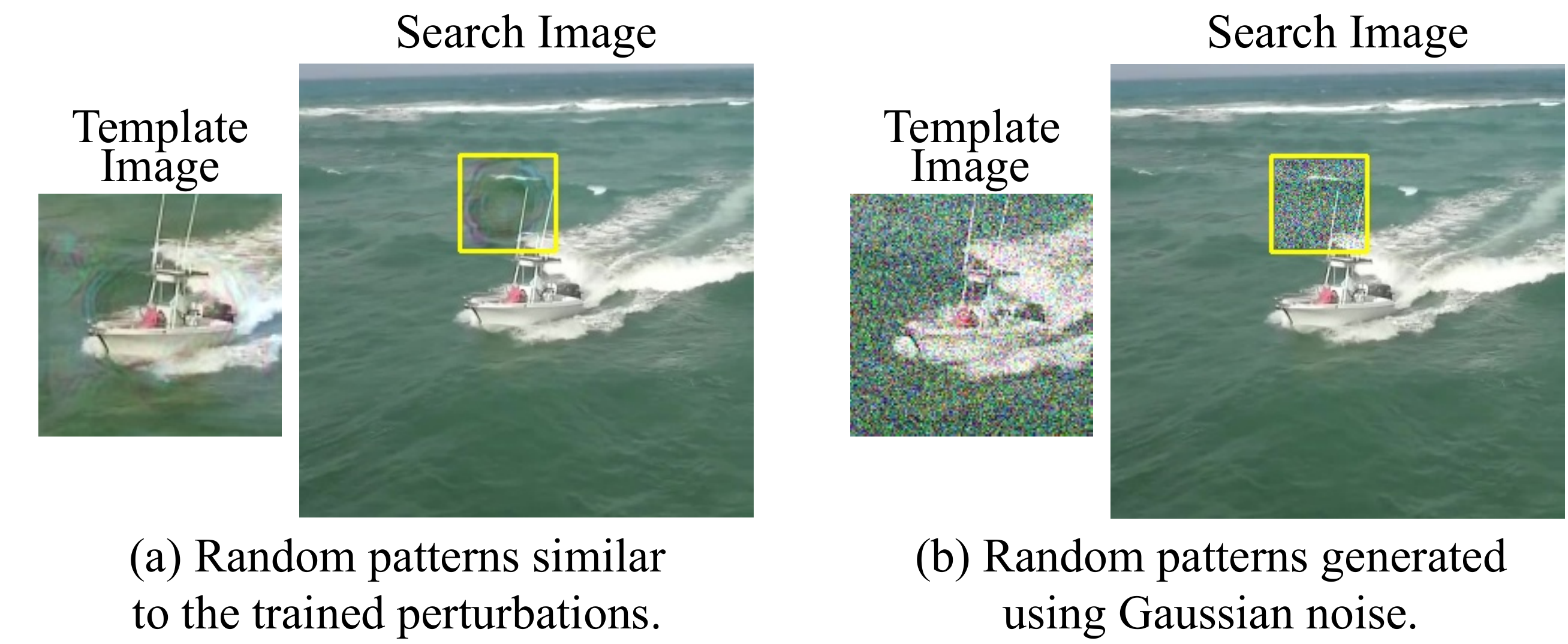}
  \caption{Visualization of the manually generated similar random patterns.}
  \label{fig:random}
\end{figure}
\begin{table}[t]
  \centering
  \caption{Attack performance comparison when only adding the perturbation on the search/template image. It is evaluated on GOT-Val.}
  \label{table:one_branch}
  \begin{tabular}{@{}cccccc@{}}
  \toprule
  \multirow{2}{*}[-2pt]{Template} & \multirow{2}{*}[-2pt]{Search} & \multicolumn{2}{c}{Untargeted Attack} & \multicolumn{2}{c}{Targeted Attack} \\ \cmidrule{3-6}
                                  &                               & AO                & SR                & AO               & SR               \\ \midrule
  \checkmark                      &                               & 0.510             & 0.567             & 0.156            & 0.106            \\
                                  & \checkmark                    & 0.714             & 0.841             & 0.160            & 0.132            \\
  \checkmark                      & \checkmark                    & 0.153             & 0.123             & 0.840            & 0.890            \\
  \bottomrule
  \end{tabular}
\end{table}
\begin{table}[!ht]
  \centering
  \caption{Analysis of the impact of each loss component on GOT-Val.}
  \begin{tabular}{ccccccc} 
  \toprule
  \multirow{2}{*}[-2pt]{$L_{\text{cls}}$}     & \multirow{2}{*}[-2pt]{$L_{\text{quality}}$} & \multirow{2}{*}[-2pt]{$L_{\text{reg}}$} & \multicolumn{2}{c}{Targeted Attack}          & \multicolumn{2}{c}{Untargeted Attack}           \\ 
  \cmidrule{4-7}
                         &                    &                    & AO                    & SR                    & AO                    & SR                     \\ 
  \midrule
  \checkmark   &    &    & 0.643  & 0.726    & 0.282 & 0.274   \\
     & \checkmark   &    & 0.148  & 0.110    & 0.747 & 0.882   \\
     &    & \checkmark   & 0.726  & 0.762    & 0.243 & 0.238   \\
  \checkmark   & \checkmark   & \checkmark   & 0.840  & 0.890    & 0.153 & 0.123   \\ \bottomrule
  \end{tabular}
  \vspace{-3mm}
  \label{tab:loss}
\end{table}
\begin{table}[t]
  \centering
  \caption{Attack results on OTB2015, GOT-Val and LaSOT with or without the \uline{ground-truth} information.}
  \begin{tabular}{rrcccc}
  \toprule
  \multirow{2}{*}[-2pt]{Benchmarks} & \multirow{2}{*}[-2pt]{Metrics} & \multicolumn{2}{c}{Untargeted Attack} & \multicolumn{2}{c}{Targeted Attack} \\ \cmidrule{3-6}
                              &                          & w/ GT  & \multicolumn{1}{l}{w/o GT}  & w/ GT  & \multicolumn{1}{l}{w/o GT} \\ \midrule
  \multirow{2}{*}{OTB2015}     & AO                       & 0.063  & 0.056                       & 0.759  & 0.752                      \\
                              & Precision                & 0.092  & 0.080                       & 0.795  & 0.794                      \\ \midrule
  \multirow{2}{*}{GOT-Val}    & SR                       & 0.123  & 0.121                       & 0.890  & 0.893                      \\
                              & AO                       & 0.153  & 0.160                       & 0.840  & 0.833                      \\ \midrule
  \multirow{3}{*}{LaSOT}      & Precision                & 0.046  & 0.043                       & 0.605  & 0.531                      \\
                              & Norm. Prec.              & 0.048  & 0.044                       & 0.702  & 0.660                      \\
                              & AO                       & 0.069  & 0.063                       & 0.691  & 0.646                      \\ \bottomrule
  \end{tabular}
  \label{tab:agent_GT}
\end{table}
\begin{table}[t]
  \centering
  \caption{Influence of two different ways to generate the fake trajectory: following a fixed direction in each video and following the real trajectory. It is evaluated on GOT-Val.}
  \begin{tabular}{@{}rcccc@{}}
  \toprule
  \multirow{2}{*}[-2pt]{Type of the Fake Trajectory} & \multicolumn{2}{c}{Untargeted Attack} & \multicolumn{2}{c}{Targeted Attack} \\ \cmidrule{2-5}
                              & AO                & SR                & AO               & SR               \\ \midrule
  Fixed direction             & 0.175             & 0.144             & 0.845            & 0.897            \\
  Follow the real trajectory  & 0.153             & 0.123             & 0.840            & 0.890            \\ \bottomrule        
  \end{tabular}
  \vspace{-3mm}
  \label{table:direction}
\end{table}
\begin{table}[t]
  \centering
  \caption{Transferability to different backbones of our perturbations trained on SiamFC++\_GoogleNet. It is evaluated on GOT-Val.}
  \begin{tabular}{rcccccc} 
  \toprule
  \multirow{2}{*}[-2pt]{Backbones} & \multicolumn{2}{c}{Before Attack} & \multicolumn{2}{c}{Untargeted Attack} & \multicolumn{2}{c}{Targeted Attack}  \\ 
  \cmidrule{2-7}
                            & AO    & SR                           & AO    & SR                           & AO    & SR                           \\ 
  \midrule
  GoogLeNet                 & 0.760 & 0.897                        & 0.153 & 0.123                        & 0.840 & 0.890                        \\
  AlexNet                   & 0.720 & 0.850                        & 0.496 & 0.577                        & 0.327 & 0.336                        \\
  ShuffleNet                & 0.766 & 0.888                        & 0.496 & 0.557                        & 0.409 & 0.426                       \\
  \bottomrule
  \end{tabular}
  \label{tab:backbone}
\end{table}
\begin{table}[t]
  \centering
  \caption{Transferability to different tracking architectures of our perturbations trained on SiamFC++\_GoogleNet. It is evaluated on OTB2015.}
  \begin{tabular}{rcccc} 
  \toprule
  \multirow{2}{*}[-2pt]{Trackers} & \multicolumn{2}{c}{Before Attack} & \multicolumn{2}{c}{Untargeted Attack}  \\
  \cmidrule{2-5}
                            & AO & Precision              & AO & Precision                   \\
  \midrule
  SiamRPN++~\cite{SiamRPN++}               & 0.676   & 0.879                  & 0.418   & 0.556                       \\
  SiamRPN~\cite{SiamRPN}                  & 0.666   & 0.876                  & 0.483   & 0.643                       \\
  Ocean~\cite{zhang2020ocean}                 & 0.672   & 0.902             & 0.237 & 0.282     \\ \bottomrule
  \end{tabular}
  \vspace{-3mm}
  \label{tab:arch}
\end{table}
\begin{table*}[t]
  \centering
  \caption{State-of-the-art comparison of untargeted attack performance on VOT2016/2018 in terms of accuracy, robustness and expected average overlap (EAO).}
  \begin{tabular}{rrrcccccc}
  \toprule
  \multirow{2}{*}[-2pt]{Dataset} & \multirow{2}{*}[-2pt]{Method} & \multirow{2}{*}[-2pt]{Tracker} & \multicolumn{3}{c}{Before Attack} & \multicolumn{3}{c}{Untargeted Attack} \\ \cmidrule{4-9}
                           &                         &                          & Accuracy   & Robustness  & EAO    & Accuracy    & Robustness    & EAO     \\ \midrule
  \multirow{2}{*}{VOT2016} & RTAA                    & DaSiamRPN                & 0.625      & 0.224       & 0.439  & 0.521       & 1.613         & 0.078   \\
                           & Ours                    & SiamFC++                 & 0.626      & 0.144       & 0.460  & 0.393       & 9.061         & 0.007   \\ \midrule
  \multirow{5}{*}{VOT2018} & RTAA                    & DaSiamRPN                & 0.585      & 0.272       & 0.380  & 0.536       & 1.447         & 0.097   \\
                           & FAN                     & SiamFC                   & 0.503      & 0.585       & 0.188  & 0.420       & -             & -       \\
                           & CSA                     & SiamRPN                  & 0.570      & 0.440       & 0.261  & 0.430       & 1.900         & 0.076   \\
                           & TTP                     & SiamRPN++                & 0.600      & 0.320       & 0.340  & 0.520       & 7.820         & 0.014   \\
                           & Ours                    & SiamFC++                 & 0.587      & 0.183       & 0.426  & 0.342       & 8.981         & 0.007   \\ \bottomrule
  \end{tabular}
  \label{tab:sota_vot}
\end{table*}

\textit{Use Other Similar Random Patterns.} To further illustrate the effectiveness of our proposed method, we manually generate similar random patterns and add them on the template and search regions (see Fig. \ref{fig:random}) to show their attack performance. Specifically, we design two kinds of random patterns: (1) the random \uline{patterns} similar to our trained perturbations, and (2) the random \uline{patterns} generated using zero-mean Gaussian noise with standard deviation \uline{of} 50.0. We replace the trained perturbations with the above random patterns to attack SiamFC++\_GoogleNet, and experimentally evaluate them on GOT-Val. As shown in Table \ref{table:noise}, these random patterns cannot effectively attack the tracker.

\textit{Only Perturb the Template or Search Image.} To analyze the impact of $p$ and $\delta$ in our perturbations, we evaluate the attack performance when only adding perturbations on the template images or the search regions on GOT-Val. The result is shown in Table \ref{table:one_branch}. For the untargeted attack, only perturbing the template/search image leads to the AO of 0.510/0.714, while perturbing both the template and search \uline{images} leads to the AO of 0.153. For the targeted attack, only perturbing the template/search image leads to the AO of 0.156/0.160, while perturbing both the template and search \uline{images} leads to the AO of 0.840. Experimental results show that perturbing both of them can achieve better attack performance than only perturbing the template/search image.

\textit{Influence of Different Training Loss Components.} We implement a series of experiments to analyze the contribution of each loss component. In Table \ref{tab:loss}, we report the attack results on the GOT-Val dataset. The AO score of the tracker with respect to the real trajectory decreases from 0.760 to 0.747 when adversarial information is generated using only quality assessment loss, indicating that quality assessment loss can cause a slight degradation in the performance of the tracker. However, the tracker's AO score with respect to the \textit{fake trajectory} is only 0.148, indicating that the perturbations generated using quality assessment loss alone can barely cause the tracker to follow the specified trajectory. When adversarial information is generated using only classification loss, the tracker's AO score decreases from 0.760 to 0.282 with respect to the real trajectory and the AO with respect to the \textit{fake trajectory} is up to 0.643. This is because the perturbation information generated using classification loss causes the tracker to localize to the \uline{\textit{fake target}} instead of the real target's location. In conclusion, all loss terms are beneficial, and the classification/regression term is more important than the quality assessment term.

\textit{Attack Performance w/o \uline{Ground-truth} Information.} Our perturbations are trained using datasets with \uline{ground-truth} box information. However, we may only know the video content while the \uline{ground-truth} box information is not available in practice. In case of this scenario, we can first run the tracker that needs to be attacked on these videos and generate one prediction result for each frame. The predicted boxes can be considered as \uline{ground-truth} information for training the perturbations. To verify this, we run the off-the-shelf SiamFC++ tracker on videos in GOT-10k training set. The predicted bounding boxes are used to train our perturbations. As shown in Table \ref{tab:agent_GT}, it is effective to use the predicted boxes instead of \uline{ground-truth} boxes for training our perturbations, though the targeted attack performance is affected due to some ambiguous prediction results used for training.

\textit{An Alternative Way to Generate Fake Trajectory.} Besides the strategy to generate the \textit{fake trajectory} following the real trajectory in Sec.~\ref{setup}, we also consider an alternative way to generate the \textit{fake trajectory}. Specifically, the attacker forces the tracker to follow a fixed direction in each video. For \uline{each video}, we assign a random direction from 4 different directions, each of which consists of shifting the box away by $(\pm 3, \pm 3)$ pixels for each consecutive frame, corresponding to one of the four directions $45^{\circ}, -45^{\circ}, 135^{\circ}, -135^{\circ}$. The attack performance is evaluated on GOT-Val. As shown in Table \ref{table:direction}, our attack method achieves effective attacks under both of the two different \uline{\textit{fake trajectory}} generation ways.

\subsection{Transferability Analyses}

In this part, we analyze the transferability of our proposed attack method. Specifically, we directly apply our perturbations trained on SiamFC++\_GoogleNet to other tracking networks including SiamFC++\_ShuffleNet, SiamFC++\_AlexNet, SiamRPN, SiamRPN++ and Ocean.


\textit{Transferability to Different Backbones.} We evaluate the transferability of our attacks when applying the perturbations to two more different backbones of SiamFC++, \ie, ShuffleNet \cite{ShuffleNet} and AlexNet \cite{AlexNet}.
The experimental results are shown in Table \ref{tab:backbone}. In the case of SiamFC++\_AlexNet, the AO with respect to the real trajectory decreases from 0.72 to 0.496. Our perturbations also generalize well to SiamFC++\_ShuffleNet, despite the customized components such as pointwise group convolution and channel shuffle operation in ShuffleNet.

\textit{Transferability to Different Tracking Architectures.} We also evaluate the transferability of our attacks when applying the perturbations to three more state-of-the-art trackers: SiamRPN \cite{SiamRPN}, SiamRPN++ \cite{SiamRPN++} and Ocean \cite{zhang2020ocean}. SiamRPN and SiamRPN++ are anchor-based trackers, and Ocean is an anchor-free tracker. The experimental results are shown in Table \ref{tab:arch}. In the case of SiamRPN, the AO with respect to the real trajectory decreases from 0.666 to 0.483 and the performance of SiamRPN++ is decreased from 0.676 to 0.418. In the case of Ocean, the AO with respect to the real trajectory decreases from 0.902 to 0.282. The results show good transferability of our attacks to different tracking architectures, even if the generated perturbations are applied to anchor-based trackers.

\subsection{Comparison with Other Attack Methods}\label{SOTA}

We firstly compare our attack method with the recent state-of-the-art attack methods on OTB2015, including CSA \cite{CSA}, RTAA \cite{RTAA}, SPARK \cite{SPARK}, FAN \cite{FAN} and TTP \cite{TTP}. \uline{Please refer to Sec.~\ref{attacktracker} for more details about these methods. The untargeted attack results in Table \ref{tab:SOTA} show that our targeted attack method can also achieve superior untargeted attack performance as well as the RTAA, SPARK and TTP attack methods and even better results than CSA and FAN. What's more, the targeted attack results in Table \ref{tab:SOTA1} also show that our achieved precision score with respect to the \textit{fake trajectory} is up to 0.795, which is significantly better than FAN and TTP. This further demonstrates our superior targeted attack performance when using the adversarial patch for attacks.}

We also compare our untargeted attack results on VOT2016 and VOT2018 with CSA \cite{CSA}, RTAA \cite{RTAA}, FAN \cite{FAN} and TTP \cite{TTP} in Table \ref{tab:sota_vot}. In summary, compared with other attack methods, our method has two key advantages. First, the proposed attack method only needs to perform the addition operations to deploy the adversarial information for any novel video without gradient optimization or network inference, making it possible to attack a real-world online-tracking system when we can not get access to the limited computational resources. Second, the proposed perturbations show good transferability to other anchor-free or anchor-based trackers. The main limitation of our work is that the translucent perturbations \uline{may} result in suspicious attacks.

\begin{table}[t]
  \centering
  \caption{State-of-the-art comparison of untargeted attack performance on OTB2015 in terms of precision score.}
  \begin{tabular}{@{}rrccc@{}}
  \toprule
  \multirow{2}{*}[-1pt]{Method} & \multirow{2}{*}[-1pt]{Tracker} & \multirow{2}{*}[-1pt]{\begin{tabular}[c]{@{}c@{}}Attack Cost\\per Frame(ms)\end{tabular}} & \multirow{2}{*}[-1pt]{\begin{tabular}[c]{@{}c@{}}Before\\ Attack\end{tabular}} & \multirow{2}{*}[-1pt]{Untargeted Attack} \\
   &  &  &  &     \\ \midrule
  RTAA & DaSiamRPN & - & 0.880 & 0.050\\
  SPARK & SiamRPN & 41.4 & 0.851 & 0.064\\
  CSA & SiamRPN & 9 & 0.851 & 0.458\\
  FAN & SiamFC & 10 & 0.720 & 0.180\\
  TTP & SiamRPN++ & 8 & 0.910 & 0.080 \\
  \midrule
  Ours & SiamFC++ & $\sim 0$ & 0.861 & 0.092\\ \bottomrule
  \end{tabular}
  \label{tab:SOTA}
\end{table}
\begin{table}[t]
  \centering
  \caption{State-of-the-art comparison of targeted attack performance on OTB2015 in terms of precision score.}
  \begin{tabular}{@{}rrc@{}}
  \toprule
  Method & Tracker &  Targeted Attack \\
  \midrule
  FAN & SiamFC  &0.420 \\
  TTP & SiamRPN++ &0.692 \\
  \midrule
  Ours & SiamFC++  &0.795 \\ \bottomrule
  \end{tabular}
  \label{tab:SOTA1}
\end{table}

\section{Conclusion}
In this paper, we propose a video-agnostic targeted attack method for Siamese trackers. We aim to attack the tracker by adding a perturbation to the template image and adding a \textit{fake target}, \ie, a small adversarial patch, into the search image adhering to the predefined trajectory, so that the tracker outputs the location and size of the \textit{fake target} instead of the real target. Being universal, the generated perturbations can be conveniently exploited to perturb videos on-the-fly without extra computations. Extensive experiments on several popular datasets show that our method can effectively fool the Siamese trackers in a targeted attack manner. In the future work, we expect that it will be possible to further reduce the possible suspicion by increasing the perturbation's imperceptibility while maintaining the attack efficiency.

\section*{Acknowledgments}
The authors would like to thank the reviewers for their constructive comments and suggestions.

\normalem
\bibliographystyle{IEEEtran}
\bibliography{ref}

\begin{thebibliography}{10}
\providecommand{\url}[1]{#1}
\csname url@samestyle\endcsname
\providecommand{\newblock}{\relax}
\providecommand{\bibinfo}[2]{#2}
\providecommand{\BIBentrySTDinterwordspacing}{\spaceskip=0pt\relax}
\providecommand{\BIBentryALTinterwordstretchfactor}{4}
\providecommand{\BIBentryALTinterwordspacing}{\spaceskip=\fontdimen2\font plus
\BIBentryALTinterwordstretchfactor\fontdimen3\font minus
  \fontdimen4\font\relax}
\providecommand{\BIBforeignlanguage}[2]{{%
\expandafter\ifx\csname l@#1\endcsname\relax
\typeout{** WARNING: IEEEtran.bst: No hyphenation pattern has been}%
\typeout{** loaded for the language `#1'. Using the pattern for}%
\typeout{** the default language instead.}%
\else
\language=\csname l@#1\endcsname
\fi
#2}}
\providecommand{\BIBdecl}{\relax}
\BIBdecl

\bibitem{DBLP:journals/tcsv/ShanZLZH21}
Y.~Shan, X.~Zhou, S.~Liu, Y.~Zhang, and K.~Huang, ``{SiamFPN}: A deep learning
  method for accurate and real-time maritime ship tracking,'' \emph{{IEEE}
  Trans. Circuits Syst. Video Technol.}, vol.~31, no.~1, pp. 315--325, 2021.

\bibitem{DBLP:journals/tcsv/FanSZYL21}
J.~Fan, H.~Song, K.~Zhang, K.~Yang, and Q.~Liu, ``Feature alignment and
  aggregation {Siamese} networks for fast visual tracking,'' \emph{{IEEE}
  Trans. Circuits Syst. Video Technol.}, vol.~31, no.~4, pp. 1296--1307, 2021.

\bibitem{DBLP:journals/tcsv/LiPWK20}
D.~Li, F.~Porikli, G.~Wen, and Y.~Kuai, ``When correlation filters meet
  {Siamese} networks for real-time complementary tracking,'' \emph{{IEEE}
  Trans. Circuits Syst. Video Technol.}, vol.~30, no.~2, pp. 509--519, 2020.

\bibitem{9258983}
M.~Jiang, Y.~Zhao, and J.~Kong, ``\uline{Mutual learning and feature fusion
  {Siamese} networks for visual object tracking},'' \emph{IEEE Trans. Circuits
  Syst. Video Technol.}, vol.~31, no.~8, pp. 3154--3167, 2021.

\bibitem{TTP}
K.~K. Nakka and M.~Salzmann, ``Temporally-transferable perturbations:
  Efficient, one-shot adversarial attacks for online visual object trackers,''
  \emph{arXiv preprint arXiv:2012.15183}, 2020.

\bibitem{FAN}
S.~Liang, X.~Wei, S.~Yao, and X.~Cao, ``Efficient adversarial attacks for
  visual object tracking,'' in \emph{Proc. Eur. Conf. Comput. Vis.}, 2020, pp.
  34--50.

\bibitem{SPARK}
Q.~Guo, X.~Xie, F.~Juefei-Xu, L.~Ma, Z.~Li, W.~Xue, W.~Feng, and Y.~Liu,
  ``{SPARK}: Spatial-aware online incremental attack against visual tracking,''
  in \emph{Proc. Eur. Conf. Comput. Vis.}, 2020, pp. 202--219.

\bibitem{chen2020one}
X.~Chen, X.~Yan, F.~Zheng, Y.~Jiang, S.-T. Xia, Y.~Zhao, and R.~Ji, ``One-shot
  adversarial attacks on visual tracking with dual attention,'' in \emph{Proc.
  IEEE Conf. Comput. Vis. Pattern Recognit.}, 2020, pp. 10\,176--10\,185.

\bibitem{UAP}
S.-M. Moosavi-Dezfooli, A.~Fawzi, O.~Fawzi, and P.~Frossard, ``Universal
  adversarial perturbations,'' in \emph{Proc. IEEE Conf. Comput. Vis. Pattern
  Recognit.}, 2017, pp. 1765--1773.

\bibitem{SiamFC++}
Y.~Xu, Z.~Wang, Z.~Li, Y.~Yuan, and G.~Yu, ``{SiamFC++}: Towards robust and
  accurate visual tracking with target estimation guidelines,'' in \emph{Proc.
  Assoc. Adv. Artif. Intell.}, 2020, pp. 12\,549--12\,556.

\bibitem{OTB}
Y.~Wu, J.~Lim, and M.-H. Yang, ``Online object tracking: A benchmark,'' in
  \emph{Proc. IEEE Conf. Comput. Vis. Pattern Recognit.}, 2013, pp. 2411--2418.

\bibitem{GOT-10k}
L.~Huang, X.~Zhao, and K.~Huang, ``{GOT-10k}: A large high-diversity benchmark
  for generic object tracking in the wild,'' \emph{arXiv preprint
  arXiv:1810.11981}, 2018.

\bibitem{LaSOT}
H.~Fan, L.~Lin, F.~Yang, P.~Chu, G.~Deng, S.~Yu, H.~Bai, Y.~Xu, C.~Liao, and
  H.~Ling, ``{LaSOT}: A high-quality benchmark for large-scale single object
  tracking,'' in \emph{Proc. IEEE Conf. Comput. Vis. Pattern Recognit.}, 2019,
  pp. 5374--5383.

\bibitem{VOT2016}
M.~Kristan, A.~Leonardis, J.~Matas, M.~Felsberg, R.~P. Pflugfelder, L.~Cehovin,
  T.~Voj{\'{\i}}r, G.~H{\"{a}}ger, A.~Lukezic, G.~Fern{\'{a}}ndez
  \emph{et~al.}, ``The visual object tracking {VOT2016} challenge results,'' in
  \emph{Proc. Eur. Conf. Comput. Vis. Workshops}, 2016, pp. 777--823.

\bibitem{VOT2018}
M.~Kristan, A.~Leonardis, J.~Matas, M.~Felsberg, R.~P. Pflugfelder, L.~C. Zajc,
  T.~Voj{\'{\i}}r, G.~Bhat, A.~Lukezic, A.~Eldesokey \emph{et~al.},
  ``\uline{The sixth visual object tracking {VOT2018} challenge results},'' in
  \emph{Proc. Eur. Conf. Comput. Vis. Workshops}, 2018, pp. 3--53.

\bibitem{VOT2019}
M.~Kristan, A.~Berg, L.~Zheng, L.~Rout, L.~V. Gool, L.~Bertinetto,
  M.~Danelljan, M.~Dunnhofer, M.~Ni, M.~Y. Kim \emph{et~al.}, ``\uline{The
  seventh visual object tracking {VOT2019} challenge results},'' in \emph{Proc.
  IEEE Int. Conf. Comput. Vis. Workshops}, 2019, pp. 2206--2241.

\bibitem{AutoTrack}
Y.~Li, C.~Fu, F.~Ding, Z.~Huang, and G.~Lu, ``{AutoTrack}: Towards
  high-performance visual tracking for {UAV} with automatic spatio-temporal
  regularization,'' in \emph{Proc. IEEE Conf. Comput. Vis. Pattern Recognit.},
  2020, pp. 11\,920--11\,929.

\bibitem{9376997}
S.~Liu, S.~Wang, X.~Liu, A.~H. Gandomi, M.~Daneshmand, K.~Muhammad, and
  V.~H.~C. De~Albuquerque, ``Human memory update strategy: A multi-layer
  template update mechanism for remote visual monitoring,'' \emph{IEEE Trans.
  Multimedia}, vol.~23, pp. 2188--2198, 2021.

\bibitem{9132673}
S.~Liu, S.~Wang, X.~Liu, C.-T. Lin, and Z.~Lv, ``Fuzzy detection aided
  real-time and robust visual tracking under complex environments,'' \emph{IEEE
  Trans. on Fuzzy Syst.}, vol.~29, no.~1, pp. 90--102, 2021.

\bibitem{zhang2020ocean}
Z.~Zhang, H.~Peng, J.~Fu, B.~Li, and W.~Hu, ``Ocean: Object-aware anchor-free
  tracking,'' in \emph{Proc. Eur. Conf. Comput. Vis.}, 2020, pp. 771--787.

\bibitem{cui2020fully}
Y.~Cui, C.~Jiang, L.~Wang, and G.~Wu, ``Fully convolutional online tracking,''
  \emph{arXiv preprint arXiv:2004.07109}, 2020.

\bibitem{9157720}
D.~Guo, J.~Wang, Y.~Cui, Z.~Wang, and S.~Chen, ``{SiamCAR}: {Siamese} fully
  convolutional classification and regression for visual tracking,'' in
  \emph{Proc. IEEE Conf. Comput. Vis. Pattern Recognit.}, 2020, pp. 6268--6276.

\bibitem{SiamRPN}
B.~Li, J.~Yan, W.~Wu, Z.~Zhu, and X.~Hu, ``High performance visual tracking
  with {Siamese} region proposal network,'' in \emph{Proc. IEEE Conf. Comput.
  Vis. Pattern Recognit.}, 2018, pp. 8971--8980.

\bibitem{SiamRPN++}
B.~Li, W.~Wu, Q.~Wang, F.~Zhang, J.~Xing, and J.~Yan, ``{SiamRPN++}: Evolution
  of {Siamese} visual tracking with very deep networks,'' in \emph{Proc. IEEE
  Conf. Comput. Vis. Pattern Recognit.}, 2019, pp. 4282--4291.

\bibitem{9339950}
S.~M. Marvasti-Zadeh, L.~Cheng, H.~Ghanei-Yakhdan, and S.~Kasaei, ``Deep
  learning for visual tracking: A comprehensive survey,'' \emph{IEEE Trans.
  Intell. Transp. Syst.}, pp. 1--26, 2021.

\bibitem{9169672}
B.~Wang, M.~Zhao, W.~Wang, F.~Wei, Z.~Qin, and K.~Ren, ``Are you confident that
  you have successfully generated adversarial examples?'' \emph{IEEE Trans.
  Circuits Syst. Video Technol.}, vol.~31, no.~6, pp. 2089--2099, 2021.

\bibitem{intriguing}
C.~Szegedy, W.~Zaremba, I.~Sutskever, J.~Bruna, D.~Erhan, I.~Goodfellow, and
  R.~Fergus, ``Intriguing properties of neural networks,'' \emph{arXiv preprint
  arXiv:1312.6199}, 2013.

\bibitem{generating}
S.~Ren, Y.~Deng, K.~He, and W.~Che, ``Generating natural language adversarial
  examples through probability weighted word saliency,'' in \emph{Proc. Assoc.
  Comput. Ling.}, 2019, pp. 1085--1097.

\bibitem{zhang2020adversarial}
W.~E. Zhang, Q.~Z. Sheng, A.~Alhazmi, and C.~Li, ``Adversarial attacks on
  deep-learning models in natural language processing: A survey,'' \emph{ACM
  Trans. Intell. Syst. Technol.}, vol.~11, no.~3, pp. 1--41, 2020.

\bibitem{morris2020textattack}
J.~X. Morris, E.~Lifland, J.~Y. Yoo, and Y.~Qi, ``{TextAttack}: A framework for
  adversarial attacks in natural language processing,'' \emph{arXiv preprint
  arXiv:2005.05909}, 2020.

\bibitem{jin2020bert}
D.~Jin, Z.~Jin, J.~T. Zhou, and P.~Szolovits, ``\uline{Is {BERT} really robust?
  {A} strong baseline for natural language attack on text classification and
  entailment},'' in \emph{Proc. Assoc. Adv. Artif. Intell.}, 2020, pp.
  8018--8025.

\bibitem{wei2019transferable}
X.~Wei, S.~Liang, N.~Chen, and X.~Cao, ``Transferable adversarial attacks for
  image and video object detection,'' in \emph{Proc. Int. Joint Conf. Artif.
  Intell.}, 2019, pp. 954--960.

\bibitem{9343885}
L.~Xiong, X.~Han, C.-N. Yang, and Y.-Q. Shi, ``Robust reversible watermarking
  in encrypted image with secure multi-party based on lightweight
  cryptography,'' \emph{IEEE Trans. Circuits Syst. Video Technol.}, pp. 1--1,
  2021.

\bibitem{9294085}
S.~Jia, X.~Li, C.~Hu, G.~Guo, and Z.~Xu, ``\uline{{3D} face anti-spoofing with
  factorized bilinear coding},'' \emph{IEEE Trans. Circuits Syst. Video
  Technol.}, vol.~31, no.~10, pp. 4031--4045, 2021.

\bibitem{meng2019white}
L.~Meng, C.-T. Lin, T.-P. Jung, and D.~Wu, ``White-box target attack for
  {EEG}-based {BCI} regression problems,'' in \emph{Proc. Int. Conf. Neural
  Inf. Process.}, 2019, pp. 476--488.

\bibitem{cheng2018query}
M.~Cheng, T.~Le, P.-Y. Chen, J.~Yi, H.~Zhang, and C.-J. Hsieh,
  ``Query-efficient hard-label black-box attack: An optimization-based
  approach,'' \emph{arXiv preprint arXiv:1807.04457}, 2018.

\bibitem{li2019nattack}
Y.~Li, L.~Li, L.~Wang, T.~Zhang, and B.~Gong, ``{NATTACK}: Learning the
  distributions of adversarial examples for an improved black-box attack on
  deep neural networks,'' in \emph{Proc. Int. Conf. Mach. Learn.}, 2019, pp.
  3866--3876.

\bibitem{papernot2017practical}
N.~Papernot, P.~McDaniel, I.~Goodfellow, S.~Jha, Z.~B. Celik, and A.~Swami,
  ``Practical black-box attacks against machine learning,'' in \emph{Proc. ACM
  Asia Conf. Comput. Commun. Security}, 2017, pp. 506--519.

\bibitem{li2020projection}
J.~Li, R.~Ji, H.~Liu, J.~Liu, B.~Zhong, C.~Deng, and Q.~Tian, ``Projection \&
  probability-driven black-box attack,'' in \emph{Proc. IEEE Conf. Comput. Vis.
  Pattern Recognit.}, 2020, pp. 362--371.

\bibitem{kurakin2018adversarial}
A.~Kurakin, I.~Goodfellow, S.~Bengio, Y.~Dong, F.~Liao, M.~Liang, T.~Pang,
  J.~Zhu, X.~Hu, C.~Xie \emph{et~al.}, ``Adversarial attacks and defences
  competition,'' \emph{arXiv preprint arXiv:1804.00097}, 2018.

\bibitem{dai2018adversarial}
H.~Dai, H.~Li, T.~Tian, X.~Huang, L.~Wang, J.~Zhu, and L.~Song, ``Adversarial
  attack on graph structured data,'' in \emph{Proc. Int. Conf. Mach. Learn.},
  2018, pp. 1115--1124.

\bibitem{li2018second}
B.~Li, C.~Chen, W.~Wang, and L.~Carin, ``Second-order adversarial attack and
  certifiable robustness,'' \emph{arXiv preprint arXiv:1809.03113}, 2018.

\bibitem{lin2017tactics}
Y.-C. Lin, Z.-W. Hong, Y.-H. Liao, M.-L. Shih, M.-Y. Liu, and M.~Sun, ``Tactics
  of adversarial attack on deep reinforcement learning agents,'' \emph{arXiv
  preprint arXiv:1703.06748}, 2017.

\bibitem{khrulkov2018art}
V.~Khrulkov and I.~Oseledets, ``Art of singular vectors and universal
  adversarial perturbations,'' in \emph{Proc. IEEE Conf. Comput. Vis. Pattern
  Recognit.}, 2018, pp. 8562--8570.

\bibitem{mopuri2018nag}
K.~R. Mopuri, U.~Ojha, U.~Garg, and R.~V. Babu, ``{NAG}: Network for adversary
  generation,'' in \emph{Proc. IEEE Conf. Comput. Vis. Pattern Recognit.},
  2018, pp. 742--751.

\bibitem{zhang2020understanding}
C.~Zhang, P.~Benz, T.~Imtiaz, and I.~S. Kweon, ``Understanding adversarial
  examples from the mutual influence of images and perturbations,'' in
  \emph{Proc. IEEE Conf. Comput. Vis. Pattern Recognit.}, 2020, pp.
  14\,521--14\,530.

\bibitem{mopuri2018generalizable}
K.~R. Mopuri, A.~Ganeshan, and R.~V. Babu, ``Generalizable data-free objective
  for crafting universal adversarial perturbations,'' \emph{IEEE Trans. Pattern
  Anal. Mach. Intell.}, vol.~41, no.~10, pp. 2452--2465, 2018.

\bibitem{chen2018shapeshifter}
S.-T. Chen, C.~Cornelius, J.~Martin, and D.~H.~P. Chau, ``{ShapeShifter}:
  Robust physical adversarial attack on {Faster R-CNN} object detector,'' in
  \emph{Proc. Joint Eur. Conf. Mach. Learn. Knowl. Discov. Databases}, 2018,
  pp. 52--68.

\bibitem{PGD}
A.~Madry, A.~Makelov, L.~Schmidt, D.~Tsipras, and A.~Vladu, ``Towards deep
  learning models resistant to adversarial attacks,'' \emph{arXiv preprint
  arXiv:1706.06083}, 2017.

\bibitem{patch}
T.~B. Brown, D.~Man{\'e}, A.~Roy, M.~Abadi, and J.~Gilmer, ``Adversarial
  patch,'' \emph{arXiv preprint arXiv:1712.09665}, 2017.

\bibitem{karmon2018lavan}
D.~Karmon, D.~Zoran, and Y.~Goldberg, ``{LaVAN}: Localized and visible
  adversarial noise,'' in \emph{Proc. Int. Conf. Mach. Learn.}, 2018, pp.
  2507--2515.

\bibitem{PAT}
R.~R. Wiyatno and A.~Xu, ``Physical adversarial textures that fool visual
  object tracking,'' in \emph{Proc. IEEE Int. Conf. Comput. Vis.}, 2019, pp.
  4822--4831.

\bibitem{GOTURN}
D.~Held, S.~Thrun, and S.~Savarese, ``Learning to track at 100 {FPS} with deep
  regression networks,'' in \emph{Proc. Eur. Conf. Comput. Vis.}, 2016, pp.
  749--765.

\bibitem{RTAA}
S.~Jia, C.~Ma, Y.~Song, and X.~Yang, ``Robust tracking against adversarial
  attacks,'' in \emph{Proc. Eur. Conf. Comput. Vis.}, 2020, pp. 69--84.

\bibitem{CSA}
B.~Yan, D.~Wang, H.~Lu, and X.~Yang, ``{Cooling-Shrinking Attack}: Blinding the
  tracker with imperceptible noises,'' in \emph{Proc. IEEE Conf. Comput. Vis.
  Pattern Recognit.}, 2020, pp. 990--999.

\bibitem{FGSM}
I.~J. Goodfellow, J.~Shlens, and C.~Szegedy, ``Explaining and harnessing
  adversarial examples,'' \emph{arXiv preprint arXiv:1412.6572}, 2014.

\bibitem{DBLP:conf/iclr/KurakinGB17a}
A.~Kurakin, I.~J. Goodfellow, and S.~Bengio, ``Adversarial examples in the
  physical world,'' in \emph{Proc. Int. Conf. Learn. Representation}, 2017.

\bibitem{shafahi2020universal}
A.~Shafahi, M.~Najibi, Z.~Xu, J.~Dickerson, L.~S. Davis, and T.~Goldstein,
  ``\uline{Universal adversarial training},'' in \emph{Proc. Assoc. Adv. Artif.
  Intell.}, 2020, pp. 5636--5643.

\bibitem{hirano2020simple}
H.~Hirano and K.~Takemoto, ``Simple iterative method for generating targeted
  universal adversarial perturbations,'' \emph{Algorithms}, vol.~13, no.~11, p.
  268, 2020.

\bibitem{focal}
T.-Y. Lin, P.~Goyal, R.~Girshick, K.~He, and P.~Doll{\'a}r, ``Focal loss for
  dense object detection,'' in \emph{Proc. IEEE Int. Conf. Comput. Vis.}, 2017,
  pp. 2980--2988.

\bibitem{iou-loss}
J.~Yu, Y.~Jiang, Z.~Wang, Z.~Cao, and T.~Huang, ``{UnitBox}: An advanced object
  detection network,'' in \emph{Proc. ACM Multimedia}, 2016, pp. 516--520.

\bibitem{SSIM}
Z.~Wang, A.~C. Bovik, H.~R. Sheikh, and E.~P. Simoncelli, ``Image quality
  assessment: from error visibility to structural similarity,'' \emph{IEEE
  Trans. Image Process.}, vol.~13, no.~4, pp. 600--612, 2004.

\bibitem{GoogLeNet}
C.~Szegedy, W.~Liu, Y.~Jia, P.~Sermanet, S.~Reed, D.~Anguelov, D.~Erhan,
  V.~Vanhoucke, and A.~Rabinovich, ``Going deeper with convolutions,'' in
  \emph{Proc. IEEE Conf. Comput. Vis. Pattern Recognit.}, 2015, pp. 1--9.

\bibitem{COCO}
T.-Y. Lin, M.~Maire, S.~Belongie, J.~Hays, P.~Perona, D.~Ramanan,
  P.~Doll{\'a}r, and C.~L. Zitnick, ``Microsoft {COCO}: Common objects in
  context,'' in \emph{Proc. Eur. Conf. Comput. Vis.}, 2014, pp. 740--755.

\bibitem{VID}
O.~Russakovsky, J.~Deng, H.~Su, J.~Krause, S.~Satheesh, S.~Ma, Z.~Huang,
  A.~Karpathy, A.~Khosla, M.~Bernstein \emph{et~al.}, ``{ImageNet} large scale
  visual recognition challenge,'' \emph{Int. J. Comput. Vis.}, vol. 115, no.~3,
  pp. 211--252, 2015.

\bibitem{8630918}
Y.~Tan, J.~Qin, X.~Xiang, W.~Ma, W.~Pan, and N.~N. Xiong, ``A robust
  watermarking scheme in {YCbCr} color space based on channel coding,''
  \emph{IEEE Access}, vol.~7, pp. 25\,026--25\,036, 2019.

\bibitem{ShuffleNet}
X.~Zhang, X.~Zhou, M.~Lin, and J.~Sun, ``{ShuffleNet}: An extremely efficient
  convolutional neural network for mobile devices,'' in \emph{Proc. IEEE Conf.
  Comput. Vis. Pattern Recognit.}, 2018, pp. 6848--6856.

\bibitem{AlexNet}
A.~Krizhevsky, I.~Sutskever, and G.~E. Hinton, ``{ImageNet} classification with
  deep convolutional neural networks,'' in \emph{Proc. Conf. Neural Inf.
  Process. Syst.}, 2012, pp. 1097--1105.

\end{thebibliography}

\begin{IEEEbiography}[{\includegraphics[width=1in,height=1.25in,clip]{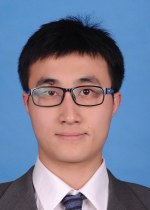}}]
{Zhenbang Li}
received the B.S. degree in computer science and technology from Beijing Institute of Technology, Beijing, China, in 2016. Currently, he is a Ph.D. student with the Institute of Automation, Chinese Academy of Sciences, Beijing, China. His research interests include object tracking and deep learning.
\end{IEEEbiography}

\begin{IEEEbiography}[{\includegraphics[width=1in,height=1.25in,clip]{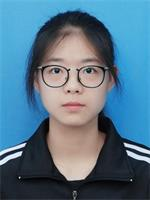}}]
  {Yaya Shi}
  received the B.S. degree from the Hefei University of Technology, Anhui, China in 2018. Currently, She is a Ph.D. student with the University of Science and Technology of China, Anhui, China. Her research interests include video captioning and deep learning.
\end{IEEEbiography}
\vspace{-5mm}

\begin{IEEEbiography}[{\includegraphics[width=1in,height=1.25in,clip]{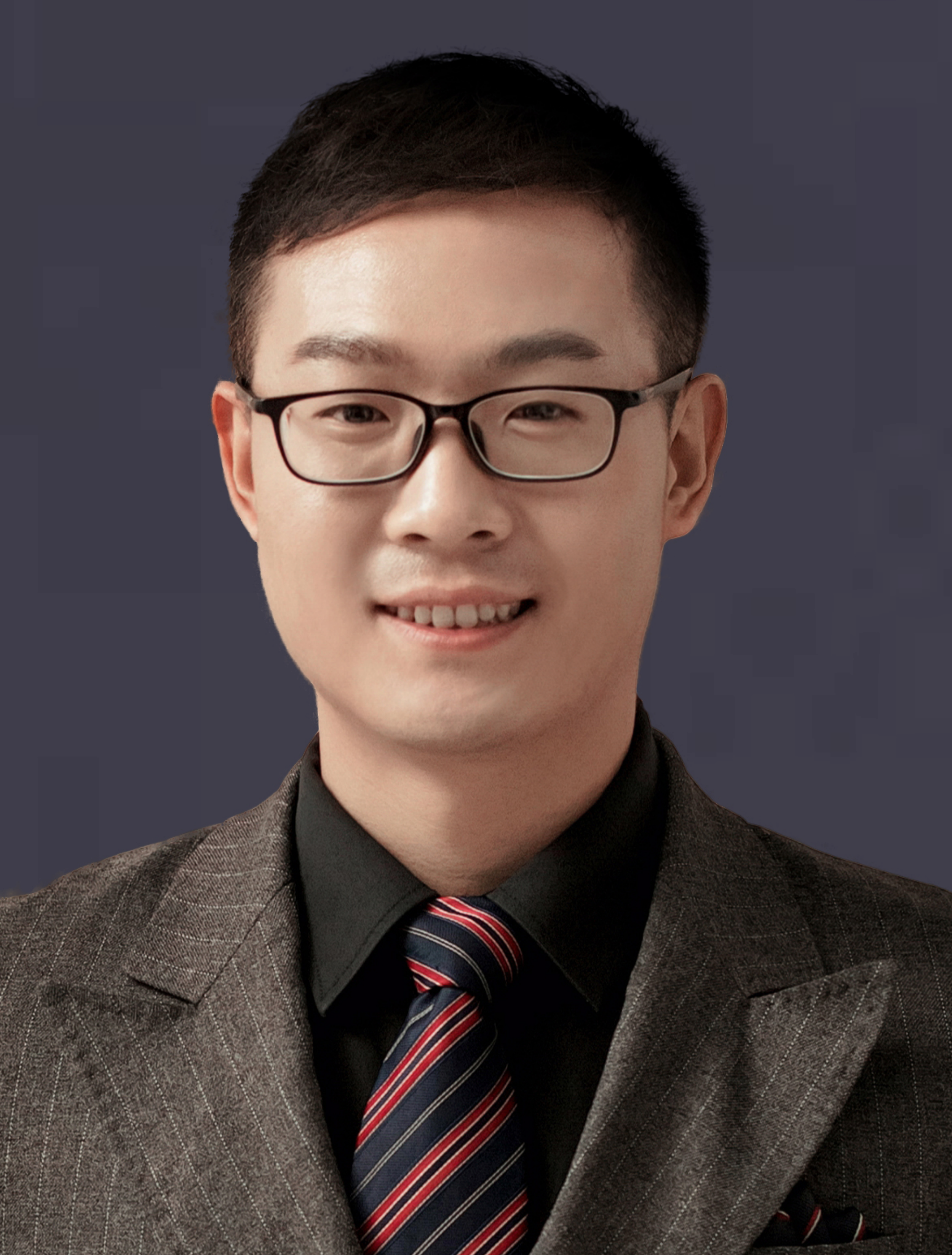}}]
{Jin Gao}
received the B.S. degree from the Beihang University, Beijing, China, in 2010, and the Ph.D. degree from the University of Chinese Academy of Sciences (UCAS), in 2015. Now he is an associate professor with the National Laboratory of Pattern Recognition (NLPR), Institute of Automation, Chinese Academy of Sciences (CASIA). His research interests include visual tracking, autonomous vehicles, and service robots.
\end{IEEEbiography}
\vspace{-5mm}

\begin{IEEEbiography}[{\includegraphics[width=1in,height=1.25in,clip]{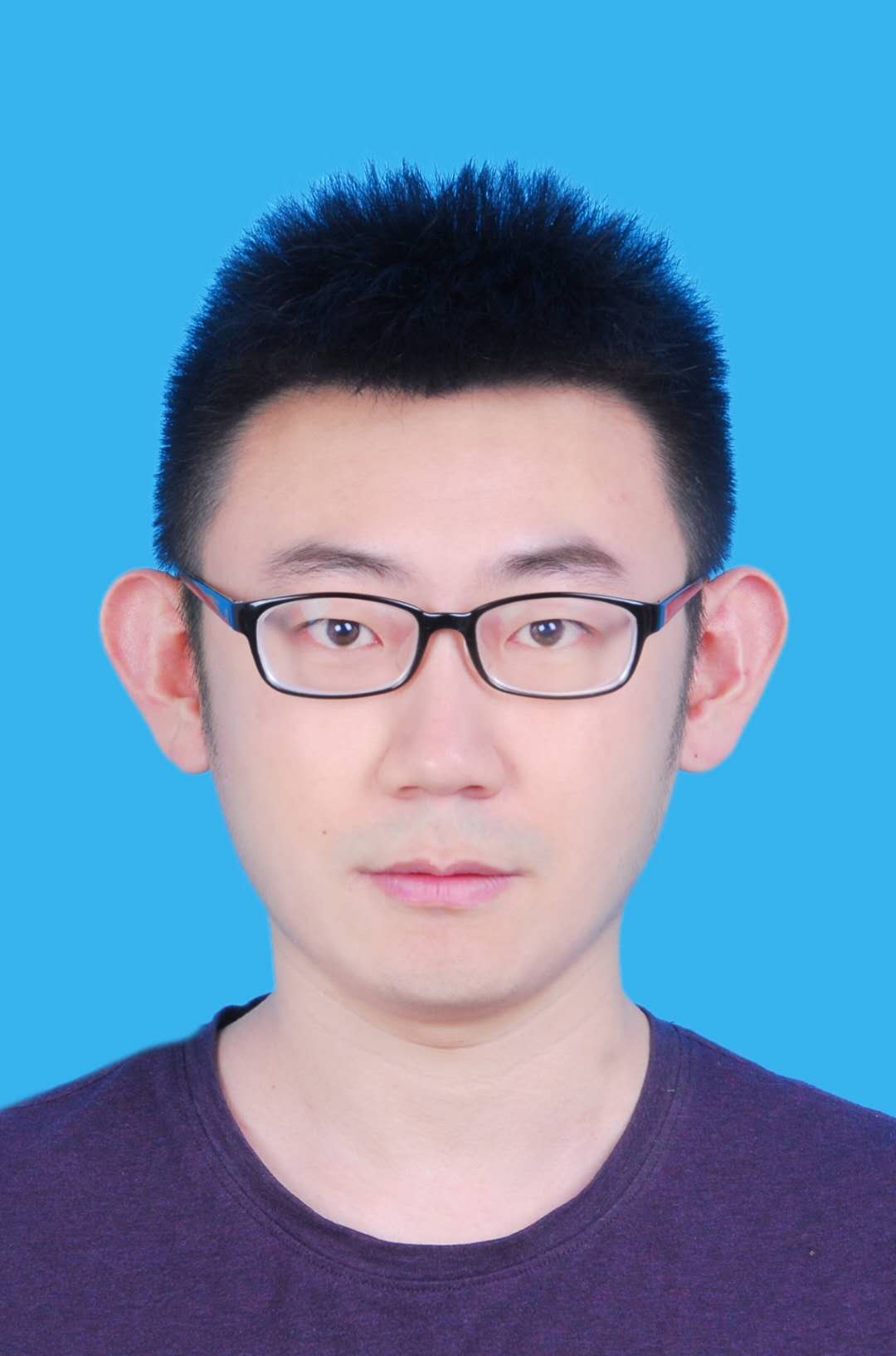}}]
{Shaoru Wang}
received the B.S. degree from the Department of School of Information and Communication Engineering in Beijing University of Posts and Telecommunications, Beijing, China in 2018. He is currently pursuing his Ph.D. degree at the Institute of Automation, Chinese Academy of Sciences (CASIA), Beijing, China. His current research interests include computer vision and deep learning.
\end{IEEEbiography}
\vspace{-5mm}

\begin{IEEEbiography}[{\includegraphics[width=1in,height=1.25in,clip]{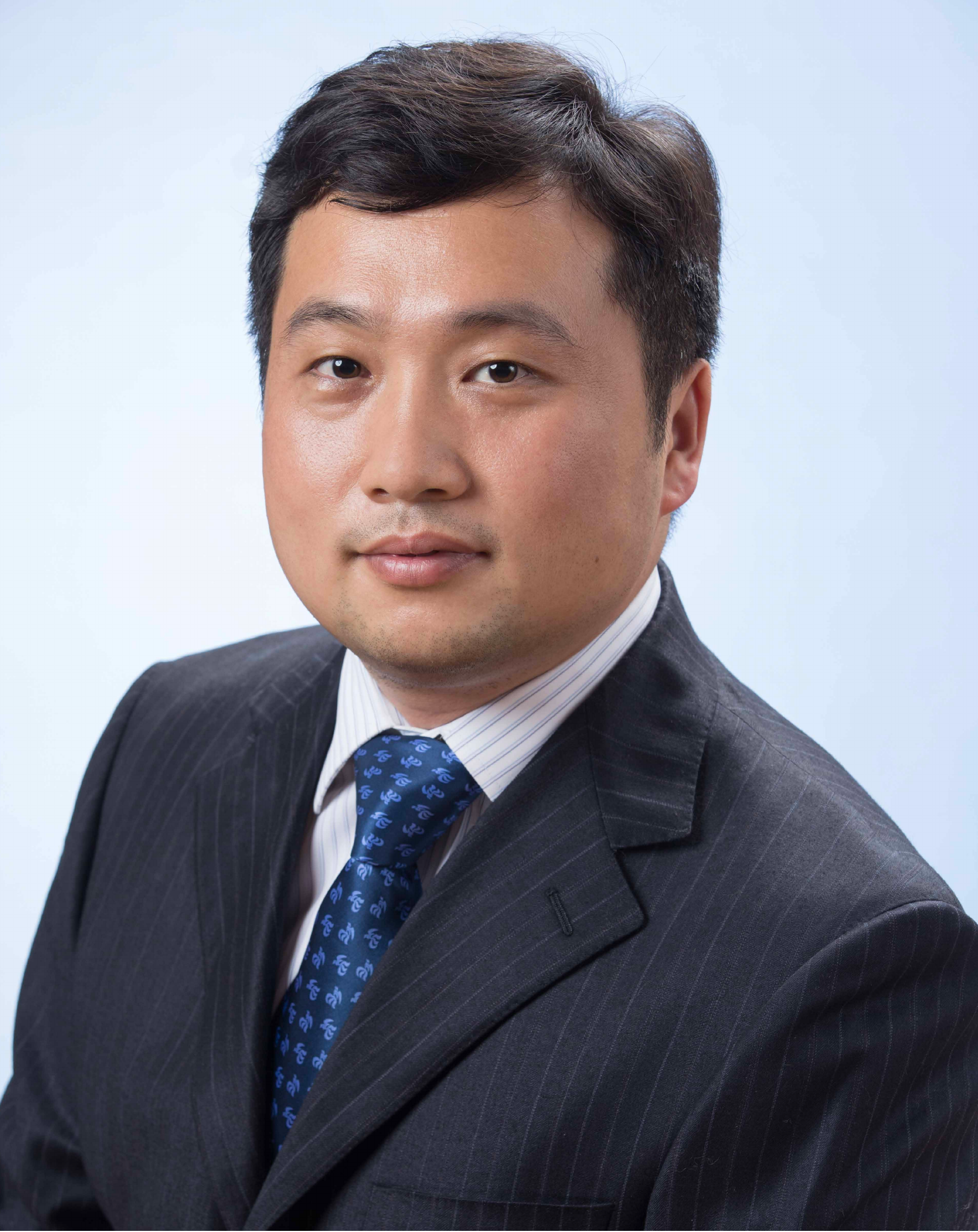}}]
{Bing Li}
received the Ph.D. degree from the Department of Computer Science and Engineering, Beijing Jiaotong University, Beijing, China, in 2009. From 2009 to 2011, he worked as a Postdoctoral Research Fellow with the National Laboratory of Pattern Recognition, Institute of Automation, Chinese Academy of Sciences (CASIA), Beijing. He is currently a Professor with CASIA. His current research interests include computer vision, color constancy, visual saliency detection, multi-instance learning, and data mining.
\end{IEEEbiography}
\vspace{-5mm}

\begin{IEEEbiography}[{\includegraphics[width=1in,height=1.25in,clip]{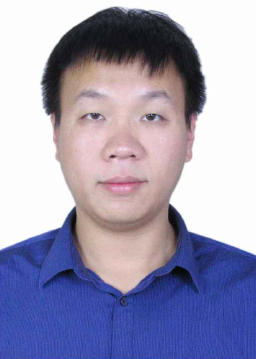}}]
{Pengpeng Liang}
received the B.S. and M.S. degrees in computer science from Zhengzhou University, China in 2008 and 2011, respectively and the Ph.D. degree from Temple University in 2016. He worked at Amazon from 2016 to 2017. Then, he joined Zhengzhou University as an assistant professor in the School of Information Engineering. His research interests include computer vision and deep learning.
\end{IEEEbiography}
\vspace{-5mm}

\begin{IEEEbiography}[{\includegraphics[width=1in,height=1.25in,clip]{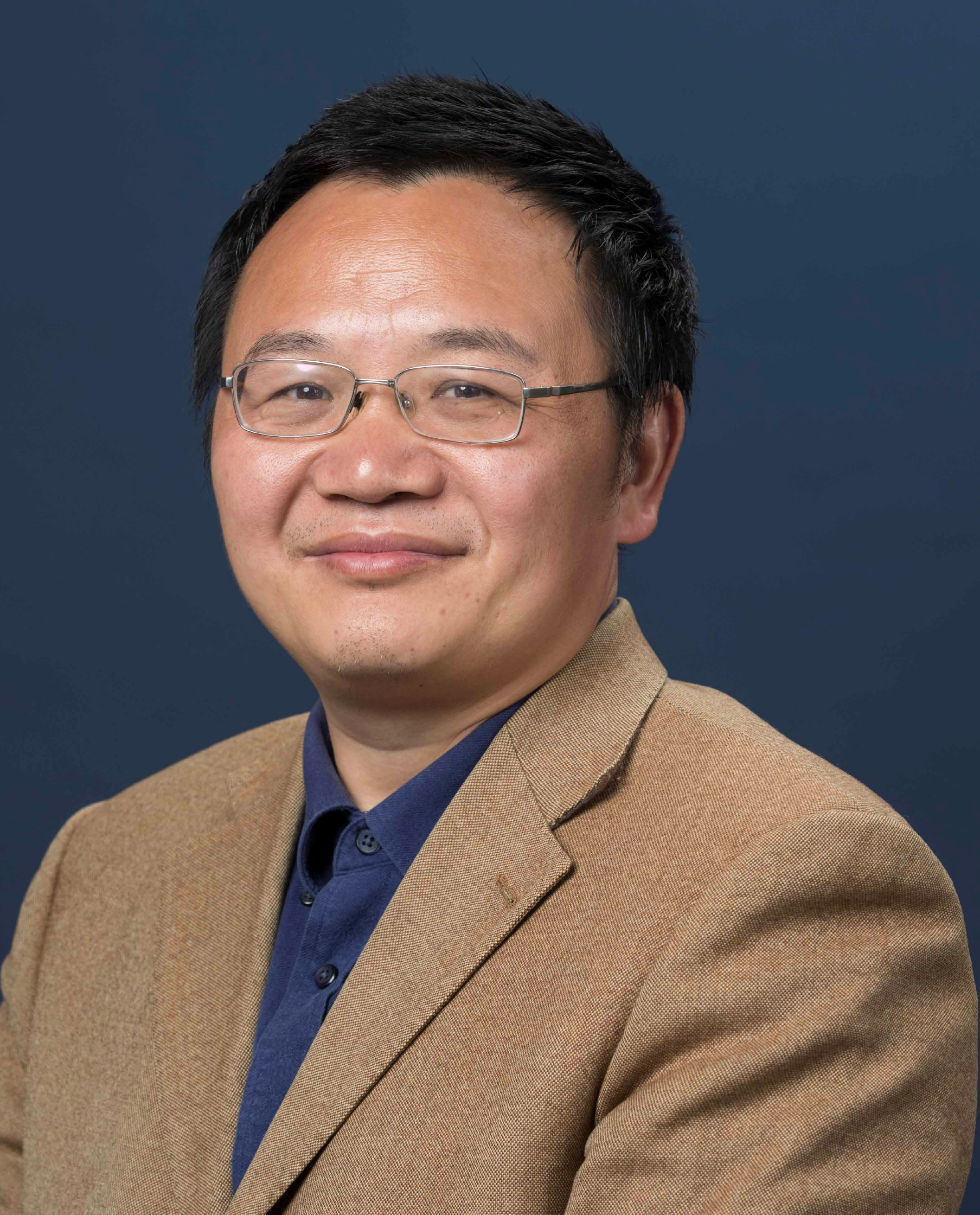}}]
{Weiming Hu}
received the Ph.D. degree from the Department of Computer Science and Engineering, Zhejiang University, Zhejiang, China. Since 1998, he has been with the Institute of Automation, Chinese Academy of Sciences (CASIA), Beijing, where he is currently a Professor. He has published more than 200 papers on peer reviewed international conferences and journals. His current research interests include visual motion analysis and recognition of harmful Internet multimedia.
\end{IEEEbiography}

\end{document}